\documentclass[11pt]{article}

\usepackage[utf8]{inputenc} 
\usepackage[T1]{fontenc}    
\usepackage{lmodern}
\usepackage{url}            
\usepackage{booktabs}       
\usepackage{amsfonts}       
\usepackage{nicefrac}       
\usepackage{microtype}      
\usepackage{float}
\usepackage{enumerate}

\usepackage{mathrsfs}
\usepackage{amsmath}
\usepackage{amssymb}
\usepackage{graphicx}
\usepackage{url}
\PassOptionsToPackage{x11names}{xcolor}
\usepackage{xcolor}
\PassOptionsToPackage{algo2e,ruled}{algorithm2e}
\usepackage{algorithm2e}
\usepackage{jmlrutils}
\usepackage{hyperref}
\usepackage{nameref}
\newcommand{\vect}{\mathrm{vec}}
\usepackage{authblk}
\DeclareMathOperator*{\argmin}{arg\,min}
\DeclareMathOperator{\R}{\mathbb{R}}

\usepackage{pdfpages} 

\begin{document}
	
\includepdf{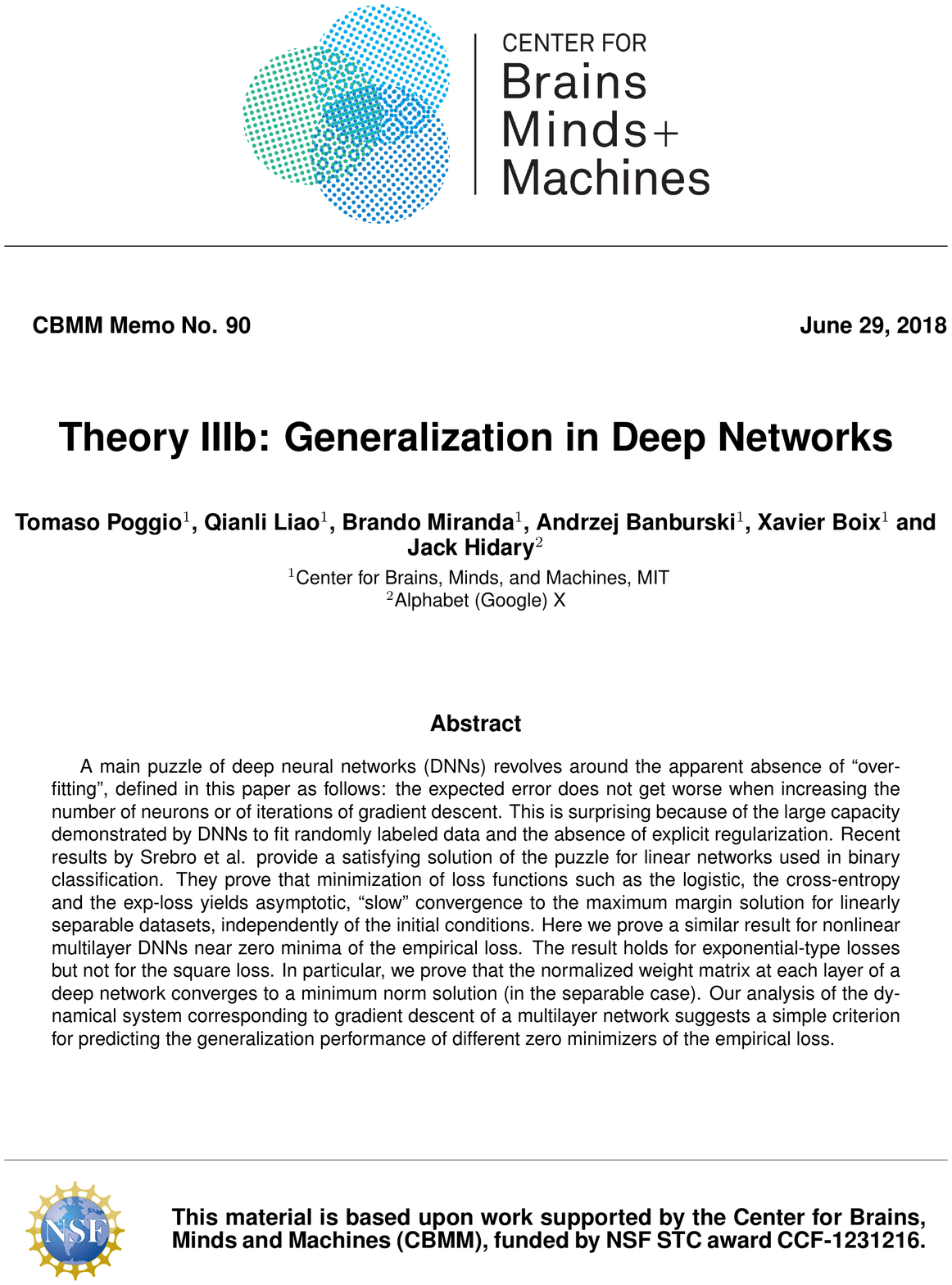} 
\setcounter{page}{1}
	
\title{Theory IIIb: Generalization in  Deep Networks}
\author{Tomaso Poggio \footnote{To whom correspondence should be addressed}}
\author {Qianli Liao}
\author {Brando Miranda}
\author{Andrzej Banburski}
\author {Xavier Boix}
\affil[1]{Center for Brains, Minds and Machines, MIT}
\author{Jack Hidary}
\affil{Alphabet (Google) X}

%

	\maketitle
	
	\begin{abstract}

          {A main puzzle of deep neural networks (DNNs) revolves
            around the apparent absence of ``overfitting'', defined in
            this paper as follows: the expected error does not get
            worse when increasing the number of neurons or of
            iterations of gradient descent. This is surprising because
            of the large capacity demonstrated by DNNs to fit randomly
            labeled data and the absence of explicit
            regularization. Recent results by
            \cite{2017arXiv171010345S} provide a satisfying solution
            of the puzzle for linear networks used in binary
            classification.  They prove that minimization of loss
            functions such as the logistic, the cross-entropy and the
            exp-loss yields asymptotic, ``slow'' convergence to the
            maximum margin solution for linearly separable datasets,
            independently of the initial conditions. Here we prove a
            similar result for nonlinear multilayer DNNs near zero
            minima of the empirical loss.  The result holds for
            exponential-type losses but not for the square loss. In
            particular, we prove that the weight matrix at
            each layer of a deep network converges to a minimum norm
            solution up to a scale factor (in the separable case). Our analysis of the
            dynamical system corresponding to gradient descent of a
            multilayer network suggests a simple criterion for
            ranking the generalization performance of different
            zero minimizers of the empirical loss.  }
		
	\end{abstract}
		
	\section{Introduction}
	
	In the last few years, deep learning has been tremendously
        successful in many important applications of machine
        learning. However, our theoretical understanding of deep
        learning, and thus the ability of developing principled
        improvements, has lagged behind. A satisfactory theoretical
        characterization of deep learning is emerging. It covers the
        following questions: 1) {\it representation power} of deep
        networks 2) {\it optimization} of the empirical risk and 3)
        {\it generalization} --- why the expected error does not
        suffer, despite the absence of explicit regularization, when
        the networks are overparametrized?  

        This paper addresses the third question which we call the
        no-overfitting puzzle, around which several recent papers
        revolve (see among others \cite{Hardt2016,
          NeyshaburSrebro2017, Sapiro2017, 2017arXiv170608498B,
          Musings2017}). We show that generalization properties of
        linear networks described in \cite{2017arXiv171010345S} and
        \cite{RosascoRecht2017} -- namely that linear networks with
        certain exponential losses trained with gradient descent
        converge to the max margin solution, providing implicit
        regularization -- can be extended to DNNs and thus resolve the
        puzzle.  We also show how the same theory can predict
        generalization of different zero minimizers of the empirical
        risk.

	\section{Overfitting Puzzle}

	\begin{figure*}[t!]\centering
		\includegraphics[width=1.0\textwidth]{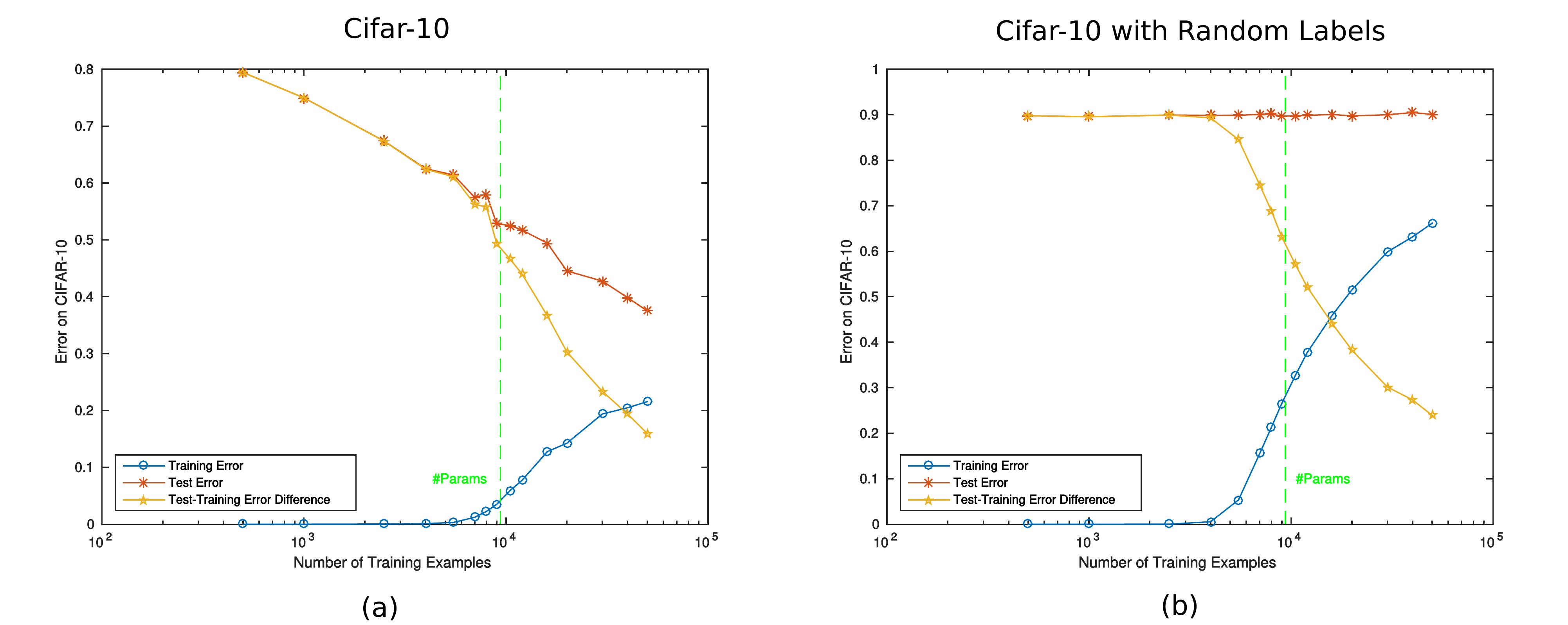}
		\caption{ {\it Generalization for Different number of Training
				Examples.} (a) Generalization error in CIFAR and (b)
			generalization error in CIFAR with random labels. The DNN was
			trained by minimizing the cross-entropy loss and it is a 5-layer
			convolutional network ({\it i.e.}, no pooling) with 16 channels per
			hidden layer. ReLU are used as the non-linearities between
			layers. The resulting architecture has approximately $10000$
			parameters.  SGD was used with batch size $= 100$ for $70$ epochs
			for each point. Neither data augmentation nor regularization is
			performed.}
		\label{TwoRegimes}
	\end{figure*}
	
	Classical learning theory characterizes generalization behavior of a
	learning system as a function of the number of training examples
	$n$. From this point of view DNNs behave as expected: the more
	training data, the smaller the test error, as shown in
	Figure~\ref{TwoRegimes}a. Other aspects of this learning curve seem
	less intuitive but are also easy to explain, {\it e.g.} the test error
	decreases for increasing $n$ even when the training error is zero (as
	noted in \cite{2017arXiv171010345S}, this is because the
	classification error is reported, rather than the risk minimized during
	training, e.g. cross-entropy). It seems that DNNs may show {\it
		generalization}, technically defined as convergence for
	$n \to \infty$ of the training error to the expected error.
	Figure~\ref{TwoRegimes} suggests generalization for increasing $n$,
	for both normal and random labels. This is expected from previous
	results such as in \cite{AntBartlett2002} and especially from the
	stability results by \cite{DBLP:journals/corr/HardtRS15}. Note that
	the property of generalization is not trivial: algorithms such as
	one-nearest-neighbor do not have this guarantee.

\begin{figure*}[t!]\centering
		\includegraphics[width=1.0\textwidth]{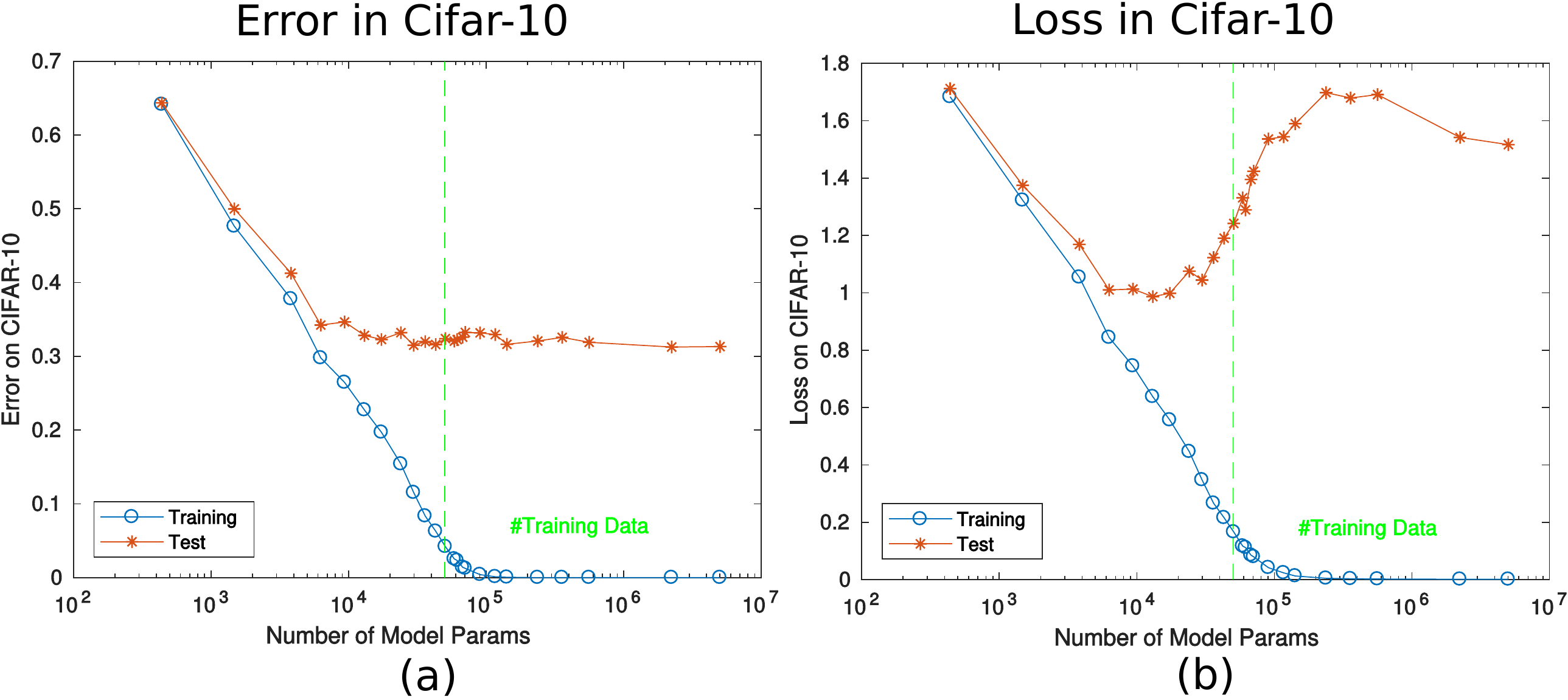}
		\caption{ {\it Expected error in CIFAR-10 as a
                    function of number of neurons.} The DNN is the
                  same as in Figure~\ref{TwoRegimes}. (a) Dependence
                  of the expected error as the number of parameters
                  increases. (b) Dependence of the cross-entropy risk
                  as the number of parameters increases.  There is
                  some ``overfitting'' in the expected risk, though
                  the peculiarities of the exponential loss function
                  exaggerate it. The overfitting in the expected loss
                  is small because SGD converges to a network with
                  minimum norm Frobenius norm for each layer (see
                  theory in the text). As a result the expected
                  classification error does not increase here when
                  increasing the number of parameters, because the
                  classification error is more robust than the loss
                  (see Appendix \ref{lackoverfitting}).
                }
		\label{Corrige:GreatPlot}
	\end{figure*}

	The property of generalization, though important, is
        of academic importance here. The real puzzle in the
        overparametrized regime typical for today's deep networks --
        and the focus of this paper-- is the apparent lack of
        overfitting in the absence of regularization. The same network
        which achieves zero training error for randomly labeled data
        (Figure \ref{TwoRegimes}b), clearly showing large capacity,
        does not show an increase in expected error when the number of
        neurons is increased in each layer without changing the
        multilayer architecture (see
        Figure~\ref{Corrige:GreatPlot}a). In particular, {\it the
          unregularized classification error on the test set does not
          get worse when the number of parameters increases well
          beyond the size of the training set}.
	
	It should be clear that the number of parameters is just a
        rough guideline to overparametrization. For details of the
        experimental setup, see Section \ref{Experiments}.

\section{Deep networks: definitions}

We define a deep network with $K$ layers with the usual
coordinate-wise scalar activation functions
$\sigma(z):\quad \mathbf{R} \to \mathbf{R}$ as the set of functions
$f(W;x) = \sigma (W^K \sigma (W^{K-1} \cdots \sigma (W^1 x)))$, where
the input is $x \in \mathbf{R}^d$, the weights are given by the
matrices $W^k$, one per layer, with matching dimensions. We use the
symbol $W$ as a shorthand for the set of $W^k$ matrices
$k=1,\cdots,K$. For
simplicity we consider here the case of binary classification in which
$f$ takes scalar values, implying that the last layer matrix $W^K$ is
$W^K \in \mathbf{R}^{1,K_l}$.
The activation functions we discuss are  the ReLU activation, linear
activation and polynomial activation.

For RELU activations the following positive homogeneity property holds
$\sigma(z)=\frac{\partial \sigma(z)}{\partial z} z$. For the network
this implies $f(W;x)=\prod_{k=1}^K \rho_k \tilde{f}(V_1,\cdots,V_K;
x_n)$, where $W_k=\rho_k V_k$ with the Frobnius norm $||V_k||=1$. In
addition, Lemma 2.1 of \cite{DBLP:journals/corr/abs-1711-01530} holds: 
\begin{equation}
\sum_{i,j} \frac{\partial f(x)}{\partial (W_k)_{i,j}}
(W_k)_{i,j}=f(x).
\label{Lemma 2.1}
\end{equation}

\section{Linear networks and dynamical systems}
	
Though we are mostly interested in the cross-entropy loss, our
analysis is applicable to the square loss and the family of losses with
exponential tails (see \cite{2017arXiv171010345S}), which include the
exponential, logistic and cross-entropy losses. For simplicity we will
mostly discuss here the simplest of them, the exponential loss, though the results follow
for the whole class. The exponential loss is of the following form:

	\begin{equation}
	L(w) = \sum_{n=1}^N \ell(y_n, f(W;x_n)).
\label{Loss}
	\end{equation}

The square loss corresponds to $\ell(y_n, f(W;x_n))=(y_n-f(W; x_n))^2$
 and the exponential loss to $\ell(y_n, f(W;x_n))=e^{-y_n f(W; x_n)}$
 with $y_n=\pm 1$ (binary classification).

 Training a network by using gradient descent is equivalent to running the
 discrete version of the gradient dynamical system defined by

	\begin{equation}
	\dot{W} = - \nabla_W L (W)= F(W)
\label{Gradient Dynamical System}
	\end{equation}
        We consider  the continuous case and therefore neglect
        the time-dependent learning rate parameter (see remarks in the
        Supplementary Material).

        In the case of one-layer, linear models -- $f(W;x)=w^Tx$ where $W^1=w^T$
        -- an explanation for the lack of overfitting has been
        recently proposed in \cite{2017arXiv171010345S}.  Two main
        properties are suggested to be important: the implicit
        regularization properties of gradient descent methods and the
        difference between classification error and the empirical loss
        which is actually minimized.  Gradient descent iteratively
        controls the complexity of the model. As the number of
        iterations can be considered the inverse of a virtual
        regularization parameter, less regularization is enforced (see
        Appendix in \cite{Theory_III}) as the number of iterations
        increase.  This description is valid for several different
        loss functions but the limit of zero regularization (or
        infinite iterations) depends on the loss function in a subtle
        and important way:
\begin{itemize}
	\item In the case of square
	loss the limit for $t \to \infty$ is the minimum norm solution {\it
		if} gradient descent starts with small weights.
	
	\item  In the case of the
	exponential loss -- and also for the logistic and cross-entropy loss
	-- the limit is again the minimum norm solution but now the
	convergence is independent of initial conditions.
\end{itemize}
In both cases, gradient descent does not change components of the
weights that are in the null space of the $x_n$ data. The proof holds
in the case of linear networks for a variety of loss functions and in
particular for the square loss (see Appendix in
\cite{DBLP:journals/corr/ZhangBHRV16} and Appendix 6.2.1 in
\cite{Theory_III}).  However, for the exponential losses the limit
$\lim_{t \to\infty}\frac{w(t)}{||w(t)||}$ used for classification will
be independent of the initial conditions on the weights.  In all
cases, the minimum norm solution is the maximum margin
solution. Intuitively, this ensures good expected classification error
for linearly separable problems.
        
	The results of \cite{2017arXiv171010345S} provide an
        interesting characterization in the case of losses with
        exponential tails. Lemma 1 in \cite{2017arXiv171010345S} shows that for loss functions
        such as cross-entropy, gradient descent on linear networks
        with separable data converges {\it asymptotically to the
          max-margin solution with any starting point $w_0$, while the
          norm $||w||$ diverges}. In particular, Theorem 3 in
        \cite{2017arXiv171010345S} states that the solution for
        $\beta$-smooth decreasing loss functions with tight
        exponential tail is $w(t)= \tilde{w} \log t +\rho(t)$ such
        that
	\begin{equation}
	\lim_{t \to\infty}\frac{w(t)}{||w(t)||} = \frac{\tilde{w}}{||\tilde{w}||}
	\end{equation}
	and that $\tilde{w}$ is the solution to the hard margin SVM,
        that is
        $ \tilde{w} = \argmin_{w \in {\R}^d} ||w||^2 $
       s.t.   $ \forall n \  \ w^Tx_n \ge 1 $.
	
	Furthermore, \cite{2017arXiv171010345S} proves that the
	{\it convergence to the maximum margin solution
		$\frac{\tilde{w}}{||\tilde{w}||} $ is  only
		logarithmic in the convergence of the empirical risk itself}. This
	explains why optimization of the logistic loss helps decrease the
	classification error in testing, even after the training
	classification error is zero and the empirical risk is very small, as
	in Figure \ref{TwoRegimes}. The conditions on
	the data that imply good classification accuracy are related to
	Tsybakov conditions (see \cite{Yao2007} and references
	therein).

\section{Nonlinear dynamics of Deep Networks}	

	Our main theorem provides and  extension of the results for linear networks to
        nonlinear deep networks by exploiting the qualitative theory of
        dynamical systems. There are two main steps in our proof:
        
        \begin{enumerate}[(a)]
        
	\item We show that linearization around an equilibrium point
          yields a linear system with weight matrices at each layer
          that, once normalized, converge asymptotically to a {\it
            finite limit} which is the {\it minimum norm} solution for
          that specific linearization and is {\it independent of
            initial conditions}.  The result does not extend to the
          square loss, as in this case the minimum norm solution for a
          linear network {\it depends on the initial conditions}.

	\item We prove that in the neighborhood of asymptotically
          stable minima of the training error, linearization of the
          nonlinear dynamics induced by the cross-entropy loss of a
          deep network describes its qualitative behavior. For this we use the
          classical Hartman-Grobman theorem (see Appendix). In particular, we show
          that the theorem is valid here for an arbitrarily small
          quadratic regularization term $\lambda P(W)$ and thus also in
          the limit $\lambda \to 0$.
        \end{enumerate}

%
%

        We explain the two steps here with more details in the
        Appendix.

\subsection{Linearization}
\label{linearization_step}

To be able to extend the linear results to nonlinear DNNs, we consider
the dynamical systems induced by GD and use classical tools to analyze
them.  The dynamical systems considered here are defined in terms of
the gradient of a potential (or Lyapunov) function that we identify
here as the empirical risk.  We are interested in the qualitative
behavior of the dynamical system Equation \ref{Gradient Dynamical
  System} near a stable equilibrium point $W_0$ where $F(W_0)=0$,
attained for $t \to \infty$. Note that we assume that gradient
descent has found a set of weights that separate the training data
that is $y_nf(x_n;W)>0, \quad \forall n=1, \cdots, N$. It easy to see
that under this assumption GD converges then to zero loss for
$t \to \infty$. 

One of the key ideas in stability theory is that the qualitative
behavior of an orbit under perturbations can be analyzed using the
linearization of the system near the orbit \cite{bhatia2002stability}.
Thus the first step is to linearize the system, which means
considering the Jacobian of $F$ or equivalently the Hessian of $-L$ at
$W_0$, that is
$ \mathbf{H}_{ij} = - \frac{\partial^{2} L}{\partial W_{i} \partial
  W_{j} }$ and evaluate it at the equilibrium. Then the (linearized)
dynamics of a perturbation $\delta W$ at $W_0$ is given by
	\begin{equation}
	\dot{\delta W} = H_{W_0} \delta W,
	\label{GradSysLin}
	\end{equation}
	
	\noindent where the matrix $H$ has only real
	eigenvalues since it is symmetric.
	
        In the case of the exponential loss
        $L(f(W))=\sum_{i=1}^N e^{- f(x_i;W) y_i }$ with a deep network
        $f$, the gradient dynamics induced by GD 
        is given by the $K$ matrix differential equations (see
        Appendix) for $k=1,\cdots,K$:

\begin{equation}
	\dot{W_k}  = \sum_{n=1}^N y_n \frac{\partial{f(x_n; W)}} {\partial W_k} e^{- y_n
		f(x_n;W)}  .
\label{ExpDynamics}
	\end{equation}
	
       We absorb here and later $y_n$ into $f(x_n;W)$ and assume that
       the new $f(x_n; W)$ is positive.  As in the linear network case of \cite{2017arXiv171010345S},
        the weights of layer $k$ that change under the dynamics must
        be in the vector space spanned by the
        $[\frac{\partial{f(x_n; W)}} {\partial W_k}]$ (which play the
        role of the data $x_n$ of the linear case). For
        overparametrized deep networks the situation is usually
        degenerate as reflected in the Hessian  which for
        large $t$ is negative semi-definite with several zero
        eigenvalues. The linearized dynamics of the perturbation is thus
        given by  $\dot{\delta W_k} = J (W) \delta W$, with

\begin{equation}
J(W)_{k  k'} =- \sum_{n=1}^N   e^{- y_n f(W_0;x_n)}\left. \left(
\frac{\partial{f(W;x_n )}} {{\partial W_k}} \frac{\partial{f(W;x_n)}}
{{\partial W_{k'}}} - y_n \frac{\partial^{2} f(W;x_n)}{\partial
	W_k \partial W_{k'} } \right)\right|_{W_0}.
\end{equation}
It is worth comparing this to the linear
case where the Hessian is
$- \sum_{n=1}^N (x^i_n) (x^j_n) e^{- (w^T x_n)}$.

The key point here is that linearization around an equilibrium
$W^0= W^0_1,\cdots W^0_k$ yields a set of $K$ equations for the
weights at each layer. The dynamics is hyperbolic with any small
regularization term and it converges to the minimum norm solution for
each $k$

\begin{lemma}
  Linearization of the nonlinear dynamics of the weight matrices $W_k$
  at each layer $k=1,\cdots,K$ yields a system of  equations
  in the weights $W_k=\tilde{W_k} \rho_k$ where $\tilde{W_k}$ are normalized
  $||\tilde{W_k}||=1$with the following properties:
\begin{enumerate}
\item each $W_k$ converges to the minimum norm
  solution. 
\item The convergence is independent of initial conditions.
\end{enumerate}

\end{lemma}

\subsection{Validity of linearization}

The question is whether linearization near an equilibrium provides a
valid description of the properties of the nonlinear system. If yes,
then the classical results for linear networks also apply to each
layer of a nonlinear deep network near an equilibrium.
		
The standard tool to prove that the behavior of the nonlinear
dynamical system associated with GD can be well described by its
linearization is the Hartman-Grobman theorem. In our case,
the theorem cannot be immediately applied.  For square loss, this is
because the minimum is in general degenerate for overparametrized deep
networks. For losses with exponential tails, this is because the
global minimum is only achieved at infinity. Both of these problems
can be solved by adding a regularization term $\lambda P(W_k)$ to the
equation for $\dot{W}_k$ for $k=1, \cdots, K$. The simplest case of
$P$ corresponds to weight decay that is
$\lambda P(W_k) =\lambda||W_k||^2$, that is the Frobenius norm for the
matrix $W_k$. We now show that the regularization term restore
hyperbolicy and can be {\it arbitrarily small}.

\begin{lemma}
  The dynamics of the weight matrices $W_k$ can be regularized by
  adding the term $\lambda_k ||W_k||^2$ to the loss function. Such
  regularization ensures hyperbolicity of the linearized dynamics
  around a zero minimizer of the empirical loss for {\it any}
  $\lambda_k >0$ and thus validity of the Hartman-Grobman theorem.
  The Hartman-Grobman theorem in turn implies that the nonlinear flow
  and the linearized flow are topologically conjugate. Thus both
  converge -- in the limit $\lambda_k \to 0$ -- to their minimum norm
  solution.

\end{lemma}

{\it Proof sketch}

As shown in more detail in the Appendix the regularized  nonlinear dynamics for
the weight matrix $W_k$ is
	
	\begin{equation}
	\dot{W_k}  =  \sum_{n=1}^n y_n
	\frac{\partial{f(x_n; W)}} {\partial W_k}  e^{- y_n
          f(x_n;W)} - \lambda_k W_k.
	\label{NonilnearReg}
\end{equation}

It can be seen that the dynamics is asymptotically hyperbolic since
the first r.h.s. matrix components decrease to zero because of the
exponential, while the second term provides stability around the
equilibrium. More detailed analyses involving the Hessian are in the
appendices. It is easy to check that, remarkably, hyperbolicity is
guaranteed for any value $\lambda_k>0$: smaller and smaller
$\lambda_k$ imply that the equilibrium is reached at longer and longer
times.  This in turns means that we can make statements about the
limit $\lambda\to 0$, in close analogy to a standard definition of the
pseudoinverse of a matrix. 

 Remember that two functions $f$ and $g$
are topologically conjugate if there exist an homeomorphism $h$ such
that $g=h^{-1} \circ f \circ h$. As an example, consider the functions
$f=a: X \to X$ and: $g={a'}: X' \to {X'}$, which are functions in the
vector spaces ${X}$ and $X'$ respectively, and $h: X\to {X'}$ is a
homeomorphism. Consider $a$ to be the matrix that solves the system of
equations $az=b$ in $X$ and $a'$ be the matrix that solves
$a' z'={b'}$ in the vector space $X'$. These systems are topologically
conjugate if and only if the dimensions of stable (negative
eigenvalues) and unstable (positive eigenvalues) subspaces of $X$ and
$X'$ match. The topological conjugacies are then $h_u : X_u \to X_u'$
and $h_s : X_s \to X_s'$, conjugating the flows on unstable and stable
subspaces. Then the map that conjugates the equations for $z$ and $z'$
is $h: x_u + y_u \mapsto h_u(x_u) + h_s(x_s)$. Note that if $f$ and
$g$ are topologically conjugate then the iterated systems $f^{(n)}$
and $g^{(n)}$ are topologically conjugate.

The application of the Hartman-Grobman theorem strictly requires
smooth activations.  We can satisfy this hypothesis by considering
polynomial approximations of the RELUs in the deep networks, since we
have empirically shown that they are equivalent to the standard
non-smooth RELUs in terms of performance. In addition, we conjecture
that the hypothesis of smooth activations is just a technicality due
to the necessary conditions for existence and uniqueness of solutions
to ODEs, which the Hartman-Grobman theorem assumes. Generalizing to
differential inclusions and non-smooth dynamical systems should allows
for these conditions to be satisfied in the Filippov sense
\cite{arscott1988differential}.

\subsection{Main result}

Putting together the lemmas, we obtain
        \begin{theorem} Given an exponential loss function and
          training data that are nonlinearly separable -- that is
          $\exists f(W; x_n)$ s.t.
          ${y_nf(W; x_n)} >0$ for all $x_n$ in the
          training set, yielding zero classification error -- the
          following properties hold around an
          asymptotic equilibrium:
	\begin{enumerate}
		\item the gradient flow induced by GD 
		is topologically equivalent to the linearized flow;
              \item the solution is the local (for the specific
                minimum) minimum Frobenius norm
                solution for the weight matrices at  each layer.
	\end{enumerate}
\end{theorem}

\begin{figure*}[t!]\centering
		\includegraphics[width=1.0\textwidth]{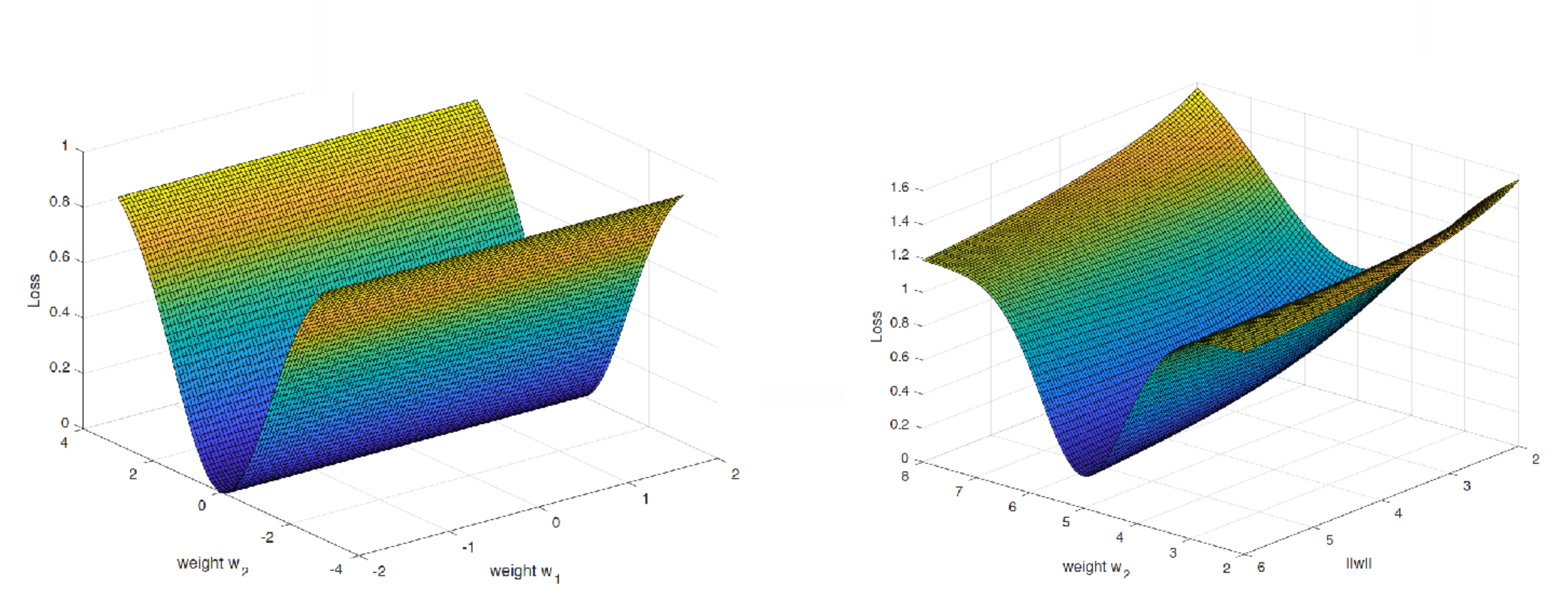}
		\caption{\it A quadratic loss function in two parameters $w_1$ and
			$w_2$ is shown on the left. The minimum has a degenerate Hessian
			with a zero eigenvalue. In the proposition described in the text, it
			represents the ``generic'' situation in a small neighborhood of zero
			minimizers with many zero eigenvalues -- and a few positive
			eigenvalues -- of the Hessian for a nonlinear multilayer network. An
			illustration of the cross-entropy risk near the global minimum at
			convergence is shown on the right part of the Figure. The valley is
			slightly sloped downwards for $||w|| \to \infty$.  In
			multilayer networks the loss function is likely to be a fractal-like
			surface with many degenerate global minima, each similar to a
			multidimensional version of the two  minima shown here. }
		\label{L2Crossentropy}
	\end{figure*}

In the {\it case of quadratic loss} the same analysis applies but
since the linearized dynamics converges to the minimum norm only for
zero initial conditions, the final statement of the theorem saying
{\it `` the solution is the local minimum norm solution''} holds only
for linear networks, such as kernel machines, but not for deep
networks. Thus the differences between the square loss and the
exponential losses becomes very significant in the nonlinear case. An
intuitive grasp of why this is, is given by Figure
\ref{L2Crossentropy}. For deep networks around a global zero minimum
the landscape of the square loss has generically many zero eigenvalues
and this is flat many directions. However, for the cross-entropy and other
exponential losses, the empirical error valleys have a small
downwards slope towards zero at infinity (see Figure
\ref{L2Crossentropy}).
	
        In the Supplementary Material we show that considering a
        related dynamics by writing $W_k = \rho_k V_k$ and imposing
        $||V_k||^2 = 1$ via a penalty parameter $\lambda$, allows us
        to show independence on initial conditions and equivalence of
        early stopping and regularization.

\subsection{Why classification is less prone to overfitting}

Because the solution is the minimum norm solution of the linearized
system, we expect, {\it for low noise data sets}\footnote{In the
  linear case this corresponds to the linear separability condition,
  while in more general settings the low noise requirement is known as
  Tsybakov conditions \cite{Yao2007}.}, little or no overfitting in
the classification error associated with minimization of the
cross-entropy \cite{2017arXiv171010345S}. Note that gradient descent
in the cross-entropy case yields convergence with linearly separable
data to the local max-margin solution {\it with any starting point}
(intuitively because of the non-zero slope in Figure
\ref{L2Crossentropy}). Thus, overfitting may not occur at all for the
expected classification error, as shown in Figure
\ref{Corrige:GreatPlot}. Usually the overfit in the associated loss is
also small, at least for almost noiseless data, because the solution
is the local maximum margin solution -- effectively the pseudoinverse
of the linearized system around the minimum.
A recent result (Corollary 2.1 in
\cite{DBLP:journals/corr/abs-1711-01530}) formally shows that the
minima of the gradient of a hinge-loss for a deep network with RELU
activations have large margin if the data are separable. The result is
consistent with our extension to nonlinear networks of the results in
\cite{2017arXiv171010345S} for exponential type losses. Note that so
far we did not make any claim about the quality of the expected
error. Different zero minimizers may have different expected errors,
though in general this rarely happen for similar initializations of
SGD.  We discuss in a separate paper how our approach here may predict
the expected error associated with each of the empirical minimizers.
	
In summary, our results imply that multilayer, deep networks behave
similarly to linear models for classification. More precisely, in the
case of {\it classification by minimization of exponential losses} the
global minimizers are guaranteed to have local maximum margin. Thus
the theory of dynamical systems suggests a satisfactory explanation of
the central {\it puzzle of non overfitting} shown in Figure
\ref{Corrige:GreatPlot}. The main result is that close to a zero
minimum of the empirical loss, the solution of the
nonlinear flow inherits the minimum norm property of the linearized
flow because the flows are topologically conjugate. Overfitting in the
loss may be controlled by regularization, explicitly (for instance via
weight decay) or implicitly (via early stopping). Overfitting in the
classification error may be avoided anyway depending on the data set,
in which case the asymptotic solution is the maximum margin solution
(for the cross-entropy loss) associated with the specific minimum.

\section{Experimental sanity check}
\label{Experiments}
	
In this paper, we focus on gradient descent (GD) rather than
stochastic gradient descent (SGD), just like the authors of
\cite{2017arXiv171010345S}. The main reason is simplicity of analysis,
since we expect the relevant results to be valid in both
cases~\cite{Theory_IIb}. In simple problems, such as in the CIFAR
dataset~\cite{CIFAR} we use in this paper, one can replace SGD with GD
without affecting the empirical results. In more difficult problems,
SGD not only converges faster but also is better at selecting global
minima {\it vs.}  local minima, for the theoretical reasons discussed
in \cite{Theory_II}. In all computer simulations shown in this paper,
we turn off all the ``tricks'' used to improve performance such as
data augmentation, weight decay, {\it etc.} This is because our goal
is to study the basic properties of DNNs optimized with gradient
descent algorithms. We keep in several figures batch normalization as it allows
to quickly reach zero training error.  We also reduce in some of the
experiments the size of the network or the size of the training
set. As a consequence, performance is not state-of-the-art, but
optimal performance is not the goal here (in fact the networks we use
achieve state-of-the-art performance using standard setups). The
expected risk was measured as usual by an out-of-sample test set.

We  test part of our theoretical analysis with the
following experiment. After convergence of GD, we apply a small random
perturbation $\delta W$ with unit norm to the parameters $W$,
then run gradient descent until the training error is again zero; this
sequence is repeated $m$ times. The dynamics of the perturbations are
given by Equation \ref{GradSysLin}. The analysis of previous sections
makes then the following predictions for the square loss:
	
	\begin{itemize}
		\item The training error will go back to zero after each sequence of
		GD.
		\item Any small perturbation of the optimum $W_0$ will be corrected by
		the GD dynamics to push back the non-degenerate weight directions to
		the original values. Since the components of the weights in
		the degenerate directions are in the null space of the gradient,
		running GD after each perturbation will not change the weights in those
		directions. Overall, the weights will change in the experiment.
		
		\item Repeated perturbations of the parameters at convergence, each
		followed by gradient descent until convergence, will not increase the
		training error but will change the parameters, increase
		norms of some of the parameters and increase the associated test error.  The $L_2$ norm
		of the projections of the weights in the null space undergoes a
		random walk  (see the Appendix).
		
	\end{itemize}
	
	The same predictions apply also to the cross entropy case with
        the caveat  that the weights increase  even without
        perturbations, though more slowly.
	Previous experiments by \cite{Theory_II} showed changes in the
	parameters and in the expected risk, consistently with our predictions above, which are
	further supported by the numerical experiments of
	Figure~\ref{CIFARclass_pert}. In the case of cross-entropy the
	almost zero error valleys of the empirical risk function are
	slightly sloped downwards towards infinity, becoming flat only
	asymptotically.

\begin{figure*}[h!]\centering
\includegraphics[width=1.0\textwidth]{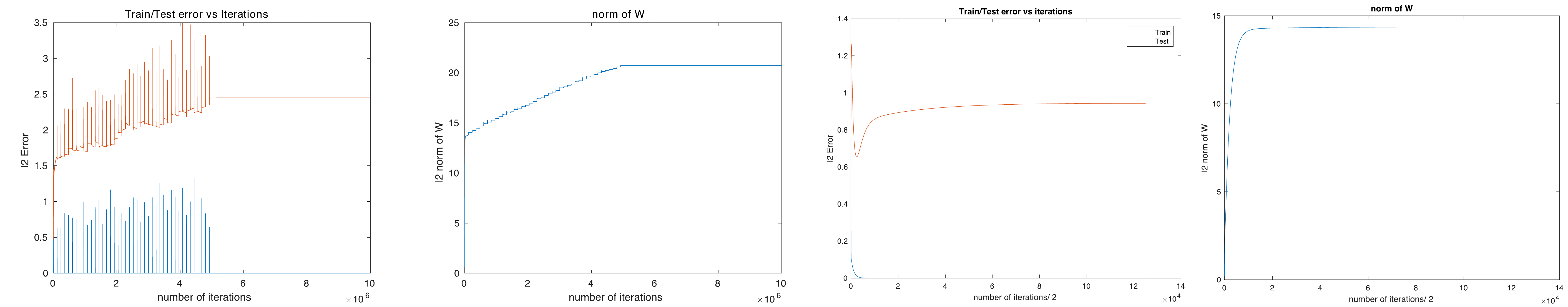}
\caption{\it Training and testing with the square loss for a linear
  network in the feature space (i.e. $y=W\Phi(X)$) with a degenerate
  Hessian of the type of Figure \ref{L2Crossentropy}. The
  target function is a sine function $f(x) = sin(2 \pi f x) $ with
  frequency $f=4$ on the interval $[-1,1]$. The number of training
  points is $9$ while the number of test points is $100$.
  For the first pair of plots the feature matrix $\phi(X)$ is a polynomial with degree 39.  
  For the first pair had points were sampled to according to the Chebyshev nodes scheme to 
  speed up training to reach zero on the train error.
  Training was done with full Gradient Descent step size $0.2$ for $10,000,000$ iterations. 
  Weights were perturbed every $120,000$ iterations and Gradient Descent was allowed to converge to zero
training error (up to machine precision) after each perturbation. The weights were perturbed
by addition of Gaussian noise with mean $0$ and standard deviation $0.45$.
The perturbation was stopped half way at iteration $5,000,000$.
The $L_2$ norm of the weights is shown in the second plot.
Note that training was repeated 29 times figures reports the average train and test error as well as average norm of the weights over the repetitions.
For the second pair of plots the feature matrix $\phi(X)$ is a polynomial with degree 30.
Training was done with full gradient descent with step size $0.2$ for $250,000$ iterations.
The $L_2$ norm of the weights is shown in the fourth plot.
Note that training was repeated 30 times figures reports the average train and test error as 
well as average norm of the weights over the repetitions.
The weights were not perturbed in this experiment.}
\label{Brando1}
\end{figure*}

	\begin{figure*}[t!]\centering
          \includegraphics[width=0.9700\textwidth]{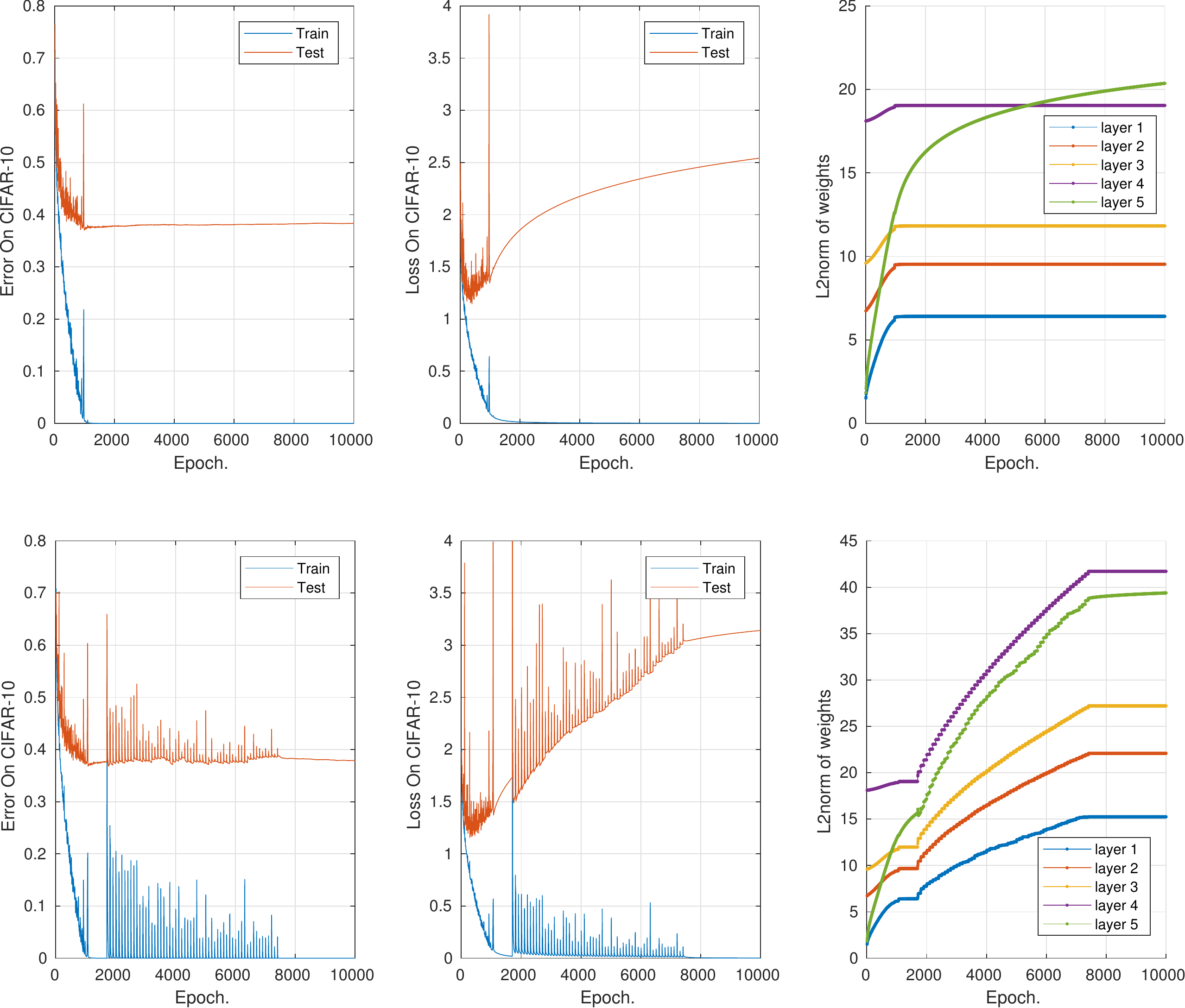}   
		\caption{\it We train a 5-layer convolutional neural
                  networks on CIFAR-10 with Gradient Descent (GD) on
                  cross-entropy loss with and without perturbations.
                  The main results are shown in the 3 subfigures in
                  the bottom row. Initially, the network was trained
                  with GD as normal. After it reaches 0 training
                  classification error (after roughly 1800 epochs of
                  GD), a perturbation is applied to the weights of
                  every layer of the network. This perturbation is a
                  Gaussian noise with standard deviation being
                  $\frac{1}{4}$ of that of the weights of the
                  corresponding layer. From this point, random
                  Gaussian noises with  such standard deviations
                  are added to every layer after every 100 training
                  epochs. The empirical risk goes back to the original
                  level after the perturbation, but the expected risk
                  grows increasingly higher. As expected, the
                  $L_2$-norm of the weights increases after each
                  perturbation step. After 7500 epochs the
                  perturbation is stopped. The left column shows the
                  classification error. The middle column shows the
                  cross-entropy risk on CIFAR during perturbations.
                  The right column is the corresponding L2 norm of the
                  weights. The 3 subfigures in the top row shows a
                  control experiment where no perturbation is
                  performed at all throughout training, The network
                  has 4 convolutional layers (filter size $3\times 3$,  
                  stride 2) and a fully-connected layer. The number of
                  feature maps (i.e., channels) in hidden layers are
                  16, 32, 64 and 128 respectively. Neither data
                  augmentation nor regularization is performed.  
                }
		\label{CIFARclass_pert}
	\end{figure*}
	
        The numerical experiments show, as predicted, that the
        behavior under small perturbations around a global minimum of
        the empirical risk for a deep networks is similar to that of
        linear degenerate regression (compare \ref{CIFARclass_pert} with Figure \ref{Brando1} ). 
        For the loss, the minimum of the expected risk may or may not occur at a finite number of
        iterations. If it does, it corresponds to an equivalent
        optimum non-zero regularization parameter $\lambda$. Thus a
        specific ``early stopping'' would be better than no
        stopping. The corresponding classification error, however, may
        not show overfitting.

\section{Putting to rest the overfitting puzzle}

Our analysis shows that deep networks, similarly to linear models,
though they may overfit somewhat the expected risk, do not usually
overfit the classification error for low-noise datasets. This follows from properties of
gradient descent for linear network, namely {\it implicit
  regularization} of the risk and the corresponding {\it margin
  maximization} for classification. In practical use of deep networks,
explicit regularization (such as weight decay) together with other
regularizing techniques (such as virtual examples) is usually added
and it is often beneficial but not  necessary, especially in the
case of classification.
	
As we discussed, the square loss is different from the exponential
loss. In the case of the square loss, regularization with arbitrarily
small $\lambda$ (in the absence of noise) restores hyperbolicity of
the gradient system and, with it, convergence to a solution. However,
the norm of the solution depends on the trajectory and is not
guaranteed to be the local minimum norm solution (in the case of
nonlinear networks) in the parameters induced by the linearization.
Without regularization, linear networks -- but not deep nonlinear networks --
are guaranteed to converge to the minimum norm solution. In the case
of the exponential loss linear networks as well as nonlinear ones
yield a hyperbolic gradient flow. Thus the solution is guaranteed to
be the maximum margin solution independently of initial conditions. For
linear networks, including kernel machines, there is a single maximum
margin solution. In the case of deep nonlinear networks there are
several maximum margin solutions, one for each of the global
minima. In some sense, our analysis shows that regularization is
mainly needed to provide hyperbolicity of the dynamics. Since this is
true also for $\lambda \to 0$ in the case of well-conditioned linear
systems, {\it the generic situation for interpolating kernel machines
  is that there is no need of regularization in the noiseless case}
(the conditioning number depends on separation of the $x$ data and is
thus independent of noise in the $y$ labels, see
\cite{2018arXiv18}). In the case of deep networks this is true only
for exponential type loss but not for the square loss.

The conclusion is that there is nothing magic in deep learning that
requires a theory different from the classical linear one with respect
to generalization, intended as convergence of the empirical to the
expected error, {\it and especially} with respect to the absence of
overfitting in the presence of overparametrization. Our analysis
explains the puzzling property of deep networks, seen in several
situations such as CIFAR, of not overfitting the expected
classification error by showing that the properties of linear networks
(for instance those emphasized by \cite{2017arXiv171010345S}) apply to
deep networks.

\section{Discussion}

Of course, the problem of establishing quantitative and useful bounds
on the performance of deep networks, remains an open and challenging
problem (see \cite{DBLP:journals/corr/abs-1711-01530}), as it is
mostly the case even for simpler one-hidden layer networks, such as
SVMs. Our main claim is that the puzzling behavior of Figure
\ref{Corrige:GreatPlot} can be explained {\it qualitatively} in terms
of the classical theory.
	
There are of a number of open problems. Though we explained the
absence of overfitting -- meant as tolerance of the expected error to
increasing number of parameters -- we did not explain in this paper why
deep networks generalize as well as they do.  In other words, this
paper explains why the test classification error in Figure
\ref{Corrige:GreatPlot} does not get worse when the number of
parameters increases well beyond the number of training data, but does
NOT explain why such test error is low.
	
We {\it conjecture} that the answer to this question may be contained
in the following theoretical framework about deep learning,
based on \cite{Theory_I}, \cite{Theory_II}, \cite{Theory_IIb},
\cite{DBLP:journals/corr/abs-1711-01530}:
	
\begin{itemize}
	\item unlike shallow networks deep networks can approximate the class
		of hierarchically local functions without incurring in the curse of
		dimensionality (\cite{MhaskarPoggio2016b,Theory_I})
       \item overparametrized deep networks yield many global
                degenerate, or almost degenerate, ``flat'' minima
                which are selected by SGD with high probability
                (\cite{Theory_IIb});
              \item overparametrization, which may yield overfit of
                the expected risk, can avoid overfitting the
                classification error for low-noise datasets because of the margin
                maximization implicitly achieved by gradient descent
                methods.
		
              \end{itemize}

              According to this framework, the main difference between
              shallow and deep networks is in terms of approximation
              power or, in equivalent words, of the ability to learn
              good representations from data based on the
              compositional structure of certain tasks. Unlike shallow
              networks, deep local networks -- in particular
              convolutional networks -- can avoid the curse of
              dimensionality  in approximating the
              class of hierarchically local compositional
              functions. This means that for such class of functions
              deep local networks represent an appropriate hypothesis
              class that allows a realizable setting, that is zero
              approximation error, with a minimum capacity.

\subsubsection*{Acknowledgments}
We thank Lorenzo Rosasco, Yuan Yao, Misha Belkin and especially Sasha
Rakhlin for illuminating
discussions. NSF funding provided by CBMM.

\newpage

\begin{center}
  {\bf APPENDICES}
\end{center}

\section{Summary: overfitting and lack of it in Figure 2}
\label{lackoverfitting}

\begin{figure*}[h!]\centering
\includegraphics[width=0.6\textwidth]{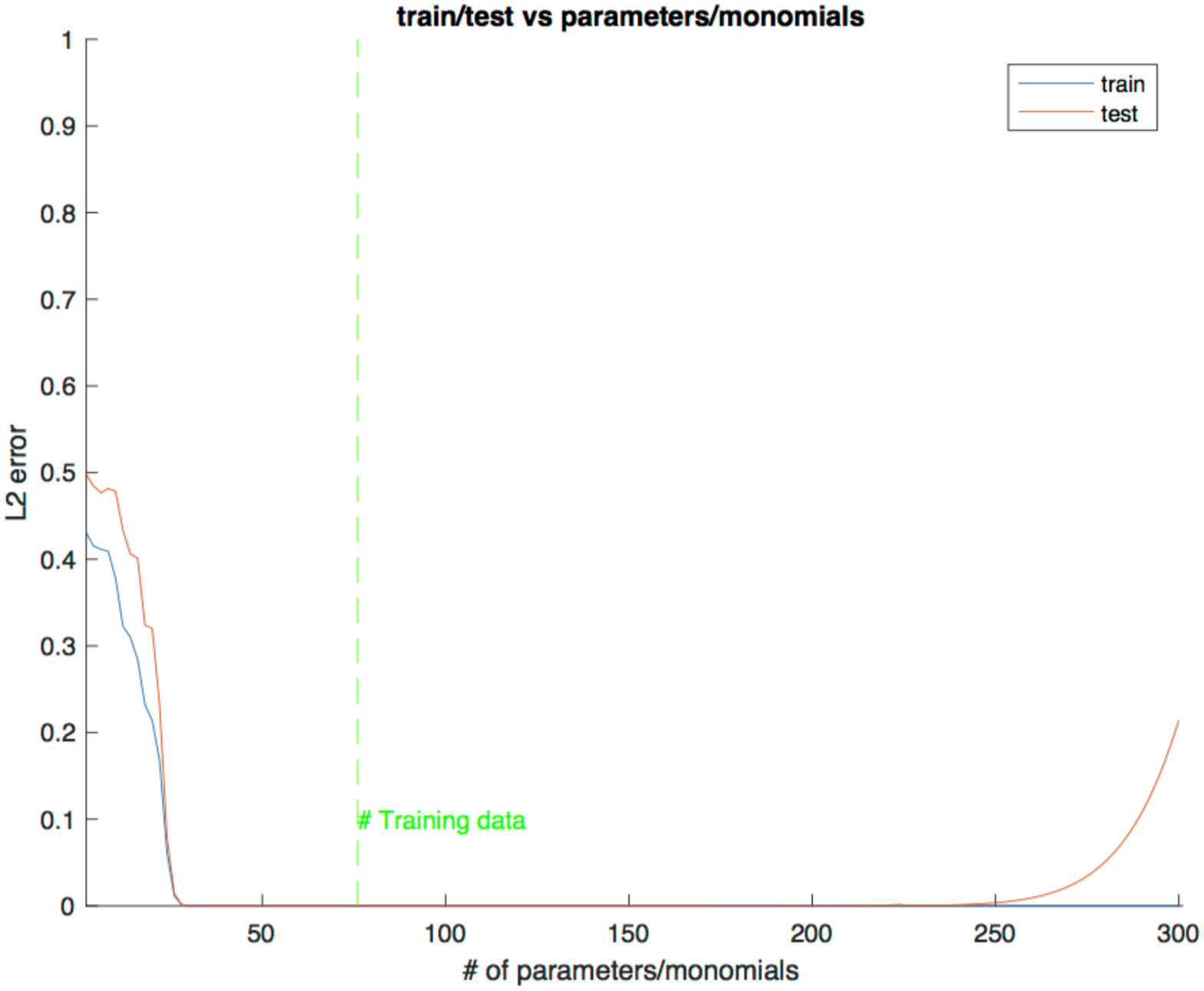}
\caption{\it Training and testing with the square loss for a linear
  network in the feature space (i.e. $y = W\phi(X)$) with a degenerate
  Hessian. The
  feature matrix is a polynomial with increasing degree, from 1 to
  300.  The square loss is plotted vs the number of monomials, that is
  the number of parameters.  The target function is a sine function
  $f(x) = sin(2 \pi f x ) $ with frequency $f=4$ on the interval
  $[-1,1]$.  The number of training points is $76$ and the number
  of test points is $600$.  The solution to the over-parametrized
  system is the minimum norm solution.  More points were sampled at
  the edges of the interval $[-1,1]$ (i.e. using Chebyshev nodes) to
  avoid exaggerated numerical errors.  The figure shows how eventually
  the minimum norm solution overfits.}
\label{Brando3}
\end{figure*}

The key reason of why there is often little or no overfitting in
overparametrized networks, such as in Figure \ref{Corrige:GreatPlot},
is that the network minimizing the training error with close to zero
loss is a minimum norm solution, as we prove in this paper in the case
of deep networks. Informally a minimum norm solution implies that the
network has the minimum complexity needed to fit the data. As an
aside, it is clear that the number of parameters is not a good measure
of the capacity or complexity of a function.  Other measures are more
appropriate such as covering numbers and entropy; norms and number of
bits are closely related. The explanation in terms of minimum norm is
classical for linear networks: the pseudoinverse solution is the best,
independently of overparametrization. It does not overfit in the ideal noiseless
case (up to numerical noise). Figure \ref{Brando3} shows an example.

This is only part of the explanation. With real data there is
always some ``noise'', either in the training or testing data, since
they do not exactly reflect the ``true'' underlying distribution. This
implies the usual appearance of  small overfitting. This is the case
for the right side of Figure \ref{Corrige:GreatPlot}. The
classification error is more resistant to
overfitting, if the data satisfy Tsybakov ``low noise'' conditions
(data density is low at the classification boundary). 
This explains the behavior
of the left side of  Figure \ref{Corrige:GreatPlot}, despite the small
overfitting of the cross-entropy loss (on the right).



\section{Hartman-Grobman theorem and dynamical systems}
\label{Hartman-Grobman theorem}

 Consider the case in which the stable point(s) of the 
	dynamical system is hyperbolic (the eigenvalues of the
	associated Hessian are negative). In this case the
	Hartman-Grobman theorem (\cite{Wanner2000}) holds (recall we assume that the
	RELUs are smoothly differentiable, since they can be replaced by
	polynomials). It says that the behavior of a dynamical system in a
	domain near a hyperbolic equilibrium point is qualitatively the same
	as the behavior of its linearization near this equilibrium point. Here
	is a version of the theorem adapted to our case.
	
	{\bf Hartman-Grobman Theorem} {\it Consider a system evolving in time as
		$\dot{w} = - F(w)$ with $F=\nabla_{w} L(w)$ a smooth map
		$F: \R^d \to \R^d$. If $F$ has a hyperbolic equilibrium state $w^*$
		and the Jacobian of $F$ at $w^*$ has no zero eigenvalues, then there
		exist a neighborhood $N$ of $w^*$ and a homeomorphism $h: N \to \R^d$,
		s.t. $h(w^*) =0$ and in $N$ the flow of $\dot{w} = - F(w)$ is
		topologically conjugate by the continuous map $U=h(w)$ to the flow of
		the linearized system $\dot{U}=-HU$ where $H$ is the Hessian of $L$.}

\subsection*{Flows}

For a linear dynamical system $\dot{x} = A x$, we can define the flow
of the solutions $\phi_t(x_o)$, which is the collection of the
solutions depending on the initial conditions. The flow is solved by
\begin{equation}
\phi_t(x_0) = e^{A t} x_0.
\end{equation}
Note, that for a symmetric $d\times d$ matrix $A$, all that really
matters for the dynamics are the eigenvalues of $A$, since we can
perform the diagonalization $A = Q \Lambda_A Q^T$, where $\Lambda_A$
is a diagonal matrix of eigenvalues of $A$ and $Q\in O(d)$ is an
orthogonal matrix. We can then write
$$
\dot{x} =  Q \Lambda_A Q^T x \Rightarrow Q^T \dot{x} = \Lambda_A Q^T x
$$
Now $Q^T$ is just a rotation or reflection in $\mathbb{R}^d$, so up to
this simple transformation, the dynamics of a linear system and its
phase portrait are governed by the eigenvalues of A.

\subsection*{Conjugacy}	
An important question in the theory of dynamical systems is whether
any two given systems are {\it different} from each other. There exists
several notions of equivalence, differing in smoothness. Here we
review three of them:

\begin{enumerate}
\item {\bf Linear conjugacy} We say that two linear systems $x' = A x$
  and $y' = B y$ are linearly conjugate iff there exists an invertible
  transformation $H$ such that $A = H^{-1} B H$ and $y = H x$. Linear
  conjugacy is thus equivalent to similarity of matrices.
	
	\item {\bf Differentiable conjugacy} For nonlinear systems, we
          can consider nonlinear changes of coordinates $y = h(x)$,
          where $h: X \to Y$ is a diffeomorphism, i.e. a continuously
          differentiable bijective map with a continuously
          differentiable inverse. We then say that an equation
          $x' = F(x)$ on some open set $\mathcal{O}_x$ is
          differentiably conjugate to $y' = G(y)$ on $\mathcal{O}_y$
          when there exists a diffeomorphism
          $h: \mathcal{O}_x\to \mathcal{O}_y$ such that the change of
          variables $y = h(x)$ converts one of the systems to the
          other. The requirement for this to happen is
	\begin{equation}
	G(y) = D_X h(h^{-1}(y))F(h^{-1}(y)).
	\end{equation}
	Around equilibria $x_{eq}$ and $y_{eq}$ of the two dynamical
        systems, the dynamics linearize to
	\begin{equation}
          u' = D_X F(x_{eq}) u \qquad \textnormal{and} \qquad v' = D_Y G(y_{eq})v
	\end{equation} 
	and the two systems are linearly conjugate by $H = D_X
        h(x_{eq})$. This implies that, like in the linear case, the
        eigenvalues of $A = D_X F(x_{eq})$ and $B = D_Y G(y_{eq})$
        have to be the same.
	
      \item{\bf Topological conjugacy} A relaxation of the above
        employs homeomorphisms (continuous bijective maps with
        continuous inverse) rather than diffeomorphisms. We say that
        two flows of dynamical systems $\phi_t : X\to X$ and
        $\psi_t : Y\to Y$ are topologically conjugate if there exists
        a homeomorphism $h: X\to Y$ such that
        $\forall x\in X \forall t\in\mathbb{R}$ we have
	\begin{equation}
	h(\phi_t(x)) = \psi_t(h(x)).
	\end{equation}
	Importantly, for linearized systems with flows $\phi_t(x) \
        e^{tA}x$ and $\psi_t(y) = e^{tB}y$ the topological conjugacy
        relaxes the statement of similarity of $A$ and $B$ to the
        requirement that the dimensions of stable and unstable spaces
        of $A$ are equal to those of $B$, i.e. only the signs of
        eigenvalues have to match.
\end{enumerate}
	
\section{Analysis: One layer linear networks}
\label{LinearAnalysis}

In this section we prove existence of a finite limit for the
normalized weight vector $\tilde{w}$ independently of initial
conditions. Our approach uses dynamical systems tools. It is more
qualitative and less detailed than \cite{2017arXiv171010345S} but it
can be used also for the nonlinear case in section \ref{Nonlinear deep
  networks}.

We consider linear networks with one layer and one scalar output
$f(W;x)=w^Tx$ with $W^1=w^T$ (multilayer linear networks have been
recently analyzed by \cite{2018arXiv180600468G}).

\subsection{Square loss} 

	Consider
\begin{equation}
	L(f(w))=\sum_{n=1}^N (y_n- w^Tx_n)^2
	\end{equation}
	
	\noindent where $y_n$ is a bounded real-valued
        variable. Assume further that the $d$-dimensional weight
        vector $w^0$ fits all the $n$ training
        data, achieving zero loss on the training set,
        that is $y_n=w^Tx_n \quad \forall n=1,\cdots,N.$

\begin{enumerate}	

\item {\it Dynamics}	The dynamics is
	
	\begin{equation}
	\dot{w} = - F(w) = - \nabla_{w} L(w)= 2 \sum_{n=1}^N E_n x_n^T
	\end{equation}
	
	\noindent with $E_n= (y_n-w^T x_n)$.
	
        The only components of the the weights that change under the
        dynamics are in the vector space spanned by the examples
        $x_n$; components of the weights in the null space of the
        matrix of examples $X^T$ are invariant to the dynamics. Thus
        $w$ converges to the minimum norm solution {\it if} the
        dynamical system starts from zero weights, as we will see also
        later.

      \item {\it Linearized dynamics} The Jacobian of $-F$ -- and Hessian of $-L$ -- for $w=w^0$
        is

	\begin{equation}
	J_{F} (w)  =  - \sum_{n=1}^N (x^i_n) (x^j_n)  
	\label{jacobian}
	\end{equation}
      
	\noindent This linearization of the dynamics around $w^0$ for
        which $L(w^0)=\epsilon_0$ yields 
	\begin{equation}
	\dot{\delta w} = J_{F} (w^0)  \delta w.
	\end{equation}
        \noindent where the associated $L$ is convex, since the
        Jacobian $J_F$ is minus the sum of auto-covariance matrices and
        thus is semi-negative definite. It is negative definite if the
        examples span the whole space but it is degenerate with some
        zero eigenvalues if $d>n$ \cite{Theory_III}.

\item {\it Regularization}	If a regularization term $\lambda w^2$ is added to the loss
        the gradient will be zero for finite values of $w$.

	In detail we have

	\begin{equation}
	\dot{w}  = -\nabla_w (L +\lambda |w|^2) =  2 \sum_{n=1}^N E_n
        x_n^T - \lambda w
	\end{equation}
	
	\noindent with 
	\begin{equation}
	J_{F} (w)  =  - \sum_{n=1}^N (x^i_n) (x^j_n)  - \lambda
	\label{jacobianReg}
	\end{equation}

\noindent which is always negative definite {\it for any arbitrarily small
$\lambda >0$}. Thus the equilibrium in
	\begin{equation}
	\dot{\delta w} = J_{F} (w^0)  \delta w.
	\end{equation}
        \noindent is hyperbolic and the Hartman-Grobman theorem
        applies.

\end{enumerate}	

In summary, regularization ensures the existence of a hyperbolic
equilibrium for any $\lambda>0$ at a finite $w^0$ (which increases to
$\infty$ for $\lambda \to 0$). If the initial conditions are $w(0)
\approx 0$, in the limit of $\lambda \to 0$ the equilibrium converges to a minimum
norm solution for $w$ and a maximum margin solution for
$\tilde{w}=\frac{w}{||w||}$. The reason is that the degenerate
directions of $w$ in which the gradient is zero will not change during
gradient descent and remain close to $0$.

\subsection{Exponential loss} 

        Consider now the exponential loss.  Even for a linear network
        the dynamical system associated with the exponential loss is
        nonlinear. While \cite{2017arXiv171010345S}  gives a rather complete characterization of
        the dynamics, here we describe a different approach based on
        linearization of the dynamics. We will then extend this
        analysis from linear networks to nonlinear networks.
	
	The exponential loss is
	
	\begin{equation}
	L(f(w))=\sum_{n=1}^N e^{- w^T x_n y_n } 
	\end{equation}
	
	\noindent where $y_n$ is a binary variable taking the value $+1$ or
	$-1$. Assume further that the $d$-dimensional weight vector
	$\tilde{w}$ separates correctly all the $n$ training data, achieving zero
	classification error on the training set, that is
	$y_i (\tilde{w})^{T} x_n \ge \epsilon, \forall n=1,\cdots,n  \quad
	\epsilon >0$. In some cases below (it will be clear from context) we incorporate $y_n$ into $x_n$. 
\begin {enumerate}		

\item {\it Dynamics}
	The dynamics is
	
	\begin{equation}
	\dot{w} = F(w) = - \nabla_{w} L(w)= \sum_{n=1}^N x_n^T e^{- x_n^T w} 
	\end{equation}
	
	\noindent thus $F(w)=\sum_{n=1}^N x_n^T e^{- x_n^T w} $.
	
	It is well-known that the weights of the networks that change
        under the dynamics must be in the vector space spanned by the
        examples $x_n$; components of the weights in the null space of
        the matrix of examples $X^T$ are invariant to the dynamics,
        exactly as in the square loss case. Unlike the square loss
        case, the dynamics of the weights diverges but the
        limit $\frac{w}{|w|}$ is finite and defines the
        classifier. This means that if a few components of the
        gradient are zero (for instance when the matrix of the
        examples is not full rank -- which is the case if $d>n$) the
        associated component of the vector $w$ will not change anymore
        and the corresponding component in $\frac{w}{|w|}$ will be
        zero. This is why there is no dependence on initial
        conditions, unlike the square loss case.
	
\item  {\it Linearized dynamics}         Though there are no equilibrium points at any finite $w$, we
        can look at the Jacobian of $F$ -- and Hessian of $-L$ -- for
        a large but finite $w=w^0$. It is
	
	\begin{equation}
	J (w)  =  - \sum_{n=1}^N (x^i_n) (x^j_n)  e^{- (w^T x_n)}
	\label{jacobianExp}
	\end{equation}
	
	\noindent The linearization of the dynamics around any finite
        $w_0$ yields a convex $L$, since $J(w_\epsilon)$ is the
        negative sum of auto-covariance matrices. The Jacobian is
        semi-negative definite in general. It is negative definite if
        the examples span the whole space but it is degenerate with
        some zero eigenvalues if $d>n$. 

The dynamics of perturbation around $w^0$ is given by
	\begin{equation}
	\dot{\delta w} = J_{F} (w^0)  \delta w.
	\end{equation}
\noindent where the degenerate directions of the gradient will be washed out asymptotically in the
        vector $\frac{w}{|w|}$ which is effectively used for classification, as
        described earlier.
	
\item   {\it Regularization}
	If an arbitrarily small regularization term such as $\lambda w^2$ is
	added to the loss, the
	gradient will be zero for finite values of $w$ -- as in the case of
	the square loss. Different components of the gradient will be zero for
	different v $w_i$. At this equilibrium point the dynamic is
	hyperbolic and the Hartman-Grobman theorem directly applies:

	\begin{equation}
	\dot{w}  = -\nabla_w (L +\lambda |w|^2) =  \sum_{n=1}^N y_n
	x^T_n  e^{- y_n (x^T_n w)}
	-\lambda w.
	\end{equation}
	
	
	
	
	The minimum is given by 
	$\sum_n \vec{x}_n e^{-x_n^T w}= \lambda \vec{w}$, which can be solved
	by $\vec{w}=\sum_n k_n \vec{x}_n$ with
	$e^{- k_n x_n\cdot \sum_j x_j} = k_n \lambda$ for $n = 1,\ldots,N$.

	The Hessian of $-L$ in the linear case for $w^0$ s.t. $\sum_n
        y_n (x_n)  e^{- y_n (x^T_n w^0)}
	=\lambda (w^0)$ is given by
	
	\begin{equation}
	-\sum_{n=1}^N 
	x^T_n x_n e^{- y_n (x^T_n w^0)}
	- \lambda 
	\end{equation}
	
	\noindent which is always negative definite, since it is the negative
	sum of the coefficients of positive semi-definite auto-covariance
	matrices and $\lambda > 0$. This means that the minimum of $L$ is
	hyperbolic and linearization gives the correct behavior for the
	nonlinear dynamical system.		
\end{enumerate}

As before for the square loss, regularization ensures the existence of
a hyperbolic equilibrium. In this case the equilibrium exists for any
$\lambda>0$ at a finite$w^0$ which increases to $\infty$ for
$\lambda \to 0$. In the limit of $\lambda \to 0$ the equilibrium
converges to a maximum margin solution for
$\tilde{w}=\frac{w}{||w||}$. The reason is that the components of $w$
in which the gradient is zero will not change during gradient
descent. Those components will be divided by a very large number (the
norm of $w$) and become zero in the normalized norm $\tilde{w}$.

\section{Analysis: Nonlinear deep networks}
	\label{Nonlinear deep networks}
\subsection{Square loss}  

	\begin{equation}
	L(f(w))=\sum_{n=1}^N (y_n-f(W;x_n))^2
	\end{equation}
	
	\noindent Here we assume that the function $f(W)$ achieves zero loss on the training set,
        that is $y_n=f(W; x_n) \quad \forall n=1,\cdots,N.$

\begin{enumerate}	
	
\item {\it Dynamics}

The dynamics now is 	
	\begin{equation}
	\dot{(W_k)_{i,j}} = - F_k(w) = - \nabla_{W_k} L(W)= 2 \sum_{n=1}^N
        E_n \frac{\partial f}{\partial (W_k)_{i,j}}
	\end{equation}
	
	\noindent with $E_n= (y_n-f(W;x_n))$.

\item  {\it Linearized dynamics} The Jacobian of $-F$ -- and Hessian of $-L$ -- for $W=W_0$ is
	
		\begin{equation}
		\begin{split}
		J (W)_{k k'}  &=  2 \sum_{n=1}^N (- (\nabla_{W_k} f(W;x_n))(\nabla_{W^{k'}}
		f(W;x_n))+E_n \nabla^2_{W_k, W^{k'}} f(W;x_n)) \\ &=- 2 \sum_{n=1}^N (\nabla_{W_k} f(W;x_n))(\nabla_{W^{k'}}
		f(W;x_n)),
		\end{split}
		\end{equation}
	
                \noindent where the last step is because of
                $E_n=0$. Note that the Hessian involves derivatives
                across different layers, which introduces interactions
                between perturbations at layers $k$ and $k'$. The
                linearization of the dynamics around $W_0$ for which
                $L(W_0)=0$ yields a convex $L$, since the Jacobian is
                semi-negative definite. In general we expect several
                zero eigenvalues because the Hessian of a deep
                overparametrized network under the square loss is
                degenerate as shown by the following theorem in
                Appendix 6.2.4 of \cite{Theory_III}:

        \begin{theorem}{\it (K. Takeuchi)} \label{thm:zero_eigenvalue}
          Let $H$ be a positive integer. Let
          $ h_k =W_{k} \sigma(h_{k-1}) \in \mathbb R^{N_k,n} $ for
          $k \in \{2,\dots,H+1\}$ and $h_1=W_{1}X$, where
          $N_{H+1}=d'$.  Consider a set of $H$-hidden layer models of
          the form, $ \hat Y_n(w) = h_{H+1}, $ parameterized by
          $w=\vect(W_1,\dots,W_{H+1})\in
          \mathbb{R}^{dN_1+N_1N_2+N_2N_3+\cdots+N_HN_{H+1}}$. Let
          $L(w)=\frac{1}{2} \|\hat Y_n(w) -Y\|^2_F$ be the objective
          function. Let $w^*$ be any twice differentiable point of $L$
          such that $L(w^*)=\frac{1}{2} \|\hat Y_n(w^*)
          -Y\|^2_F=0$. Then, if there exists $k\in \{1,\dots,H+1\}$
          such that
          $N_{k} N_{k-1} >n \cdot \min(N_{k},N_{k+1},\dots,N_{H+1})$
          where $N_0 =d$ and $N_{H+1}=d'$ (i.e., overparametrization),
          there exists a zero eigenvalue of Hessian
          $\nabla^2 L(w^{*})$.
\end{theorem}

\item  {\it Regularization}
The effect of regularization is to add the term $\lambda_k ||W_k||^2$
to the loss. This results in a Hessian of the form 

\begin{equation}
		J(W)_{k k'}  =  - 2 \sum_{n=1}^N (\nabla_{W^k} f(W;x_n))(\nabla_{W^{k'}}
		f(W;x_n)) -\lambda_k \delta_{k k'} \mathbb{I},
		\end{equation}
\end{enumerate}			

which is always negative definite for any $\lambda>0$.

\subsection{Exponential loss} 
	
	Consider again the exponential loss 
	
	\begin{equation}
	L(f(W))=\sum_{n=1}^N e^{- f(W; x_n) y_n }
\label{ExpLoss}
	\end{equation}
	
	\noindent with definitions as before. We assume that $f(W;x)$,
        parametrized by the weight vectors $W_k$, separates correctly all
        the $n$ training data $x_i$, achieving zero classification
        error on the training set for $W=W^0$, that is
        $y_if(W^0; x_n) >0, \forall n=1,\cdots,N$. Observe that if
        $f$ separates the data, then
        $\lim_{a \to\infty} L(af({W^0}))=0$ and this is where gradient
        descent converges \cite{2017arXiv171010345S}.
	
	Again there is no critical point for finite $t$. Let us
        linearize the  dynamics around a large $W^0$ by approximating  $f(W^0+\Delta W_k)$
	with a low order Taylor approximation for small $\Delta W_k$.
	
	\begin{enumerate}			
	\item {\it Dynamics}

	The gradient flow is not zero at any finite $(W^0)_k$. It is given by

\begin{equation}
	\dot{W_k}  = \sum_{n=1}^N y_n [\frac{\partial{f(W;x_n)}} {\partial W_k}]  e^{- y_n
		f(x_n;W)}  
	\end{equation}
\noindent where the partial derivatives of $f$ w.r.t. $W_k$ can be evaluated
in $W_0$.

	Let us consider a small perturbation of $W_k$ around $W^0$ in order to
	linearize $F$ around $W^0$.

\item  {\it Linearized dynamics} 

The linearized dynamics of the perturbation are $\dot{\delta W_k} = J (W)  \delta W$, with

\begin{equation}
J(W)_{k  k'} =- \sum_{n=1}^N   e^{- y_n f(W_0;x_n)}\left. \left(
\frac{\partial{f(W;x_n )}} {{\partial W_k}} \frac{\partial{f(W;x_n)}}
{{\partial W_{k'}}} - y_n \frac{\partial^{2} f(W;x_n)}{\partial
	W_k \partial W_{k'} } \right)\right|_{W^0}.
\end{equation}

Note now that the term containing the second derivative of $f$ does not
vanish at a minimum, unlike in the square loss case. Indeed, away from the minimum
this term contributes negative eigenvalues.

\item {\it Regularization}

	Adding a regularization term of the form $\sum_{i=1}^K
        \lambda_k ||W_k||^2$ yields for $i=1,\cdots,K$
	
	\begin{equation}
	\dot{W_k}  = -\nabla_w (L +\lambda |W_k|^2) =  \sum_{n=1}^n y_n
	\frac{\partial{f(W; x_n)}} {\partial W_k}  e^{- y_n   f(x_n; W)} -
	 \lambda_k W_k
\end{equation}

	For compactness of notation, let us define
	\begin{equation}
	g^{(n)}_k = y_n \frac{\partial f}{\partial W_k} e^{-y_n f(W; x_n)},
	\end{equation}
	with which we have a transcendental equation for the minimum.
	\begin{equation}
	\lambda_k ({W_k})_{min} = \sum_n g^{(n)}_k.\label{eq:transcendental_minimum}
	\end{equation}

	The Jacobian of $F$ (and negative Hessian of loss) is then
	\begin{equation}
	J(W)_{k  k'} = \sum_n \frac{\partial g^{(n)}_k}{\partial W_{k'}} - \lambda_k \delta_{k k'} \mathbb{I}.
	\end{equation}
	
\item   At this new finite equilibrium the Hessian is now positive definite
  for any $\lambda_i > 0$.  This guarantees that a perturbation
  $\delta W$ around $W_0$ remains small: there is asymptotic
  stability.  Furthermore, for the exponential loss -- but not for the
  square loss -- the dynamics for any $W$
  close to $W_0$  remains qualitatively the
  same when $\lambda \to 0$, in other words is not affected by the
  presence of regularization. The parameters resulting from
  linearization may be different from the original weights: the minimum
  norm solution is in terms of these new local parameters.


\item {\it Normalized dynamics}

We consider here the dynamics of the normalized network with
normalized weight matrices $\tilde{W_k}$ induced by the gradient
dynamics of $W_k$, where $W_k$ is the weight matrix of layer $k$. We
note that this normalized dynamics is related to the technique called
``weight normalization'' used in gradient descent\cite{SalDied16}.
For simplicity of notation we consider here for each weight matrix
$W_k$ the corresponding ``vectorized'' representation in terms of a
vector that we denote as $w$ (dropping the index $k$ for convenience).

We  use the following definitions and self-evident properties:

\begin{itemize}
\item Define $\frac{w}{||w||}=\tilde{w}$; thus $w=||w||\tilde{w}$ with
  $||\tilde{w}||=1$.
\item The following relations are easy to check:
\begin{enumerate}
\item $\frac{\partial ||w||}{\partial w}=\tilde{w}$
\item $\frac{\partial \tilde{w}}{\partial
    w}=\frac{I-ww^T}{||w||}=S$. $S$ has at most one zero eigenvalue
  since  $ww^T$ is rank $1$ with a single eigenvalue
$\lambda_1=1$. 
\item $Sw=S \tilde{w}=0$
\item $||w||S^2=S$
\item $\frac{\partial ||\tilde{w}||^2}{\partial w}=0$
\label{Relations}
\end{enumerate}

\item We assume  $f(w)=f(||w||,\tilde{w}) =||w|| f(1, \tilde{w})=||w||
  \tilde{f}$. 

\item Thus $\frac{\partial f}{\partial w}=\tilde{w} \tilde{f} +|w|
  S \frac{\partial \tilde{f}}{\partial \tilde {w}}$ 

\item The gradient descent dynamic system used in training deep networks for the
  exponential loss of Equation \ref{ExpLoss} is given by Equation
  \ref{ExpDynamics}, that is by

	\begin{equation}
	\dot{w}  = -\frac{\partial L}{\partial w}= \sum_{n=1}^N y_n \frac{\partial{f(x_n; w)}} {\partial w_i}  e^{- y_n
		f(x_n;w)} 
	\end{equation}
\noindent with a Hessian given by (assuming $y_n f(x_n) >0$)
	\begin{equation}
	H = \sum_{n=1}^N  e^{- f(x_n;w)} (\frac{\partial{f(x_n; w)}}
        {\partial w} {\frac{\partial{f(x_n; w)}} {\partial w}}^T - \frac{\partial^2{f(x_n; w)}} {\partial w^2})
	\end{equation}

\item The dynamics above for $w$ induces the following dynamics for
  $||w||$ and $\tilde{w}$:
\begin{equation}
\dot{||w||}= \frac{\partial ||w||}{\partial w} \dot{w}= \tilde{w} \dot{w}
\end{equation}
\noindent and
\begin{equation}
\dot{\tilde{w}}= \frac{\partial \tilde{w}}{\partial w} \dot{w}= S \dot{w}
\end{equation}

Thus
\begin{equation}
\dot{||w||}= \tilde{w}^T \dot{w}=\frac{1}{||w||} \sum_{n=1}^N w^T \frac{\partial{f(x_n; w)}} {\partial w_i}  e^{- 
		f(x_n;w)}= \sum_{n=1}^N  e^{-
                ||w|| \tilde{f}(x_n)} \tilde{f}(x_n)  
\label{wnormdot}
\end{equation}
\noindent where, assuming that $w$ is the vector corresponding to the
weight matrix of layer $k$, we obtain $(w^T \frac{\partial f(w;x)}{\partial w})= f(w;x)$
because of Lemma 1 in \cite{DBLP:journals/corr/abs-1711-01530}.  We
assume that $f$ separates all the data, that is
$y_n f(x_n)>0  \ \  \forall n$.  Thus $\frac{d}{dt}{||w||} >0$
and $\lim_{t \to \infty}\dot{||w||}=0$. In the 1-layer network case
the dynamics yields $||w|| \approx \log t$ asymptotically. For deeper
networks, this is different. In Section \ref{Weightgrowth} we show
that the product of weights at each layer diverges faster than
logarithmically, but each individual layer diverges slower than in the
1-layer case.  By defining
\begin{equation}
\sum_n e^{- \||w|| \tilde{f}(x_n)} \frac{\partial
  \tilde{f}(x_n)}{\partial \tilde {w}}= \tilde{B}
\end{equation}

\noindent the Equation above becomes
\begin{equation}
\dot{\tilde{w}}=\frac{I-\tilde{w}\tilde{w}^T}{||w||}\tilde{B}.
\label{Normalizedwdot}
\end{equation}

The dynamics imply $\dot{\tilde{w}} \to 0$ for $t \to \infty$, while
$||\tilde{w}||=1$.  As in the square loss case for $w$, the degenerate
components of $\dot{\tilde{w}}$ are not directly updated by the
gradient equation but unlike the square loss case, they are indirectly
updated because $||\tilde{w}||=1$.  Thus the dynamics is independent
of the initial conditions, unlike the dynamics of $w$ in the square
loss case. Note that the constraint $||\tilde{w}||=1$ is
automatically enforced by the definition of $\tilde{w}$.
\end{itemize}
\end{enumerate}

%

This section, and in particular inspection of Equations \ref{wnormdot}
and \ref{Normalizedwdot}, shows that the dynamics of the normalized
matrices at each layer converges. Adding a regularization term of the
form $\lambda ||W_k||^2$ and letting $\lambda$ go to $0$ supports the
following conjecture

\begin{proposition}
The normalized weight matrix at each layer $\tilde{W}_k$ converges to
the minimum Frobenius norm solution, independently of initial conditions.
\end{proposition}	


\subsection{Another approach to prove linearization of  a nonlinear deep
  network and its validity}

In this section we study the linearization of the deep nonlinear
networks around fixed points of GD and its relation with equivalent
linear networks. We first review the step of linearization described
in section \ref{linearization_step} for linear networks with either
square or exponential losses to study the dynamics of perturbations
$\dot{\delta W} = H \delta W$. We also review the same linearizaton
step for the deep nonlinear networks. Unlike the linear case, the
Hessian can have negative eigenvalues, and only becomes
positive-definite around the minimum.

We then 
 	\begin{itemize}

 	\item show that in the case of the square loss, the Hessian of
          a deep nonlinear network can be mapped to a linear network
          with appropriately transformed data;
        \item for the exponential losses we show that linearization of
          a deep nonlinear network yields a deep linear network.
 	
        \item show that the Hartman-Grobman theorem guarantees that
          the linearization faithfully describes the behavior of the
          DNNs near a minimum of GD.
 \end{itemize} 
 
   \noindent
  {\bf Proof sketch}
  
  \noindent
  We first regularize the square loss or a loss with an exponential
  tail $L$ and derive the
  continuous dynamical system in $\dot{W_k}= - \nabla_{W_k} L$ associated
  with gradient descent (with a fixed learning rate). We then
  linearize around the asymptotic equilibrium $w_0$ at which the
  gradient is zero, obtaining the dynamical system for perturbations
  $\delta W_k$ around $W^0$ in Equation (\ref{GradSysLin}).  At this
  point we check that the analysis available for linear networks --
  especially in the case of exponential losses -- applies to the
  linearized dynamical systems. For this, we need to understand the
  dynamics of perturbations $\delta W_k$ for both the linear networks
  and the deep nonlinear ones. 
  
  Note that the phase portraits of a dynamical system
  (\ref{GradSysLin}) depend solely on the eigenvalues of $H$.  For
  both the exponential losses and the square loss on linear networks,
  the Hessian is positive semi-definite everywhere, even without any
  regularization, with the number of distinct eigenvalues bounded by
  $N$ -- the number of training examples $x_n\in\mathbb{R}^d$. For the
  deep nonlinear networks, the Hessian can in general have negative
  eigenvalues, and only at a minimum of GD it become positive
  semi-definite. Interestingly, for the square loss at $w^*$ the
  Hessian has also at most $N$ distinct eigenvalues, since we have
$$H  =  2 \sum_{n=1}^N ( (\nabla_{W_k} f(W;x_n))(\nabla_{W_{k'}}
f(W;x_n)) - E_n \nabla^2_{W_k W_{k'}} f(W;x_n)),$$ where
$E_n = y_n - f(W;x_n)$ vanishes at a zero minimum.  Due to the higher
number of weights than in the one layer linear network case, the
Hessian is of higher dimensionality ($D\times D$) than in the linear
case ($D > d$). This implies that the linearization of the nonlinear
deep network with square loss corresponds to a linear system with
higher dimensional ``virtual'' data $x_n'\in\mathbb{R}^D$ related to
the original data by $x'_n = \nabla_W f(W;x_n)|_{W_0}$. This
construction provides a (stronger) differentiable conjugacy to a
linear network.
 
In the case of losses with exponential tails, the Hessian has
a non-vanishing additional term proportional to
$y_n \nabla_W^2 f(W;x_n)$. In the case of the exponential
loss we obtain
$$H_{exp} = \sum_{n=1}^N   e^{- y_n f(W;x_n)}(
\frac{\partial{f(W;x_n )}} {{\partial W_k}} \frac{\partial{f(W;x_n)}}
{{\partial W_{k'}}} - y_n \frac{\partial^{2} f(W;x_n)}{\partial
  W_k \partial W_{k'} } ). $$ In particular, derivatives across
different layers induce a higher number of distinct eigenvalues than
$N$. We show below that deep linear networks (with
$f(W;s_n) = W_{K} W_{k-1}\cdots W_2 W_1 x_n$) have the same behavior
for derivatives across layers (here and elsewhere we do not assume a
convolutional structure). The only difficulty is then of  two
derivatives in the same layer, $f''(W;x_n)$, which we remove by the
assumption of rectified
nonlinearities, for which the second derivative vanishes. Thus a deep
nonlinear network with an exponential loss linearizes to a deep linear
network with same loss, which also converges to the pseudo-inverse
like the shallow linear network.  Finally, the linearized system
satisfies the Hartman-Grobman theorem ({\it for any} $\lambda >0$) and
is therefore a good qualitative description of the dynamics of the
nonlinear system around the asymptotic equilibrium $W_0$.

\subsubsection{Square loss}	
Note that the Hessian of a deep network is of much higher
dimensionality $D > d$ for over-parametrized networks. However, since
the number of distinct eigenvalues of the linear and nonlinear
Hessians match (since they are both sums of outer products of training
example vectors), we can find a linear system with inputs
$x'\in\mathbb{R}^D$ with $x'_n = \nabla_W f(W;x_n)|_{W_0}$ and weights
$W'\in\mathbb{R}^{D}$ that satisfies the same linearized dynamical
system as the linearized deep network. Since we can explicitly match
the two Hessians, the dynamical system of a deep network with a square
loss around a minimum of gradient descent is {\it differentiably conjugate}
to a linear network with square loss.

\subsubsection{Exponential loss}	

The Hessian around $W^0$ for the exponential loss is quite different
from the square loss case for the same network $f$. This is because
the term
$y_n \frac{\partial^{2} f(W;x_n)}{\partial W_k \partial W_{k'} }$
cannot in general be written as an outer product of some vector. Let
us investigate two simple cases.
	
	\begin{enumerate}
        \item Consider a simple one-layer network, with an arbitrary
          smooth non-linear activation applied to it. In this case we
          have $f(W;x_n) = f(w^T x_n)$. It is easy to see that in this
          case
		\begin{equation}
		\frac{\partial^{2} f(w^T x_n)}{\partial w_{i} \partial w_{j} } = x_n^i x_n^j f''(	w^T x_n),
		\end{equation}
		which is again a simple outer product of a
                vector. Hence, the Hessian has again at
                most $N$ distinct eigenvalues, just like in the linear
                case. It is interesting to note that this simple case
                is also valid for a deep network, if we restrict
                ourselves to optimizing only one layer at a time. This
                extends the results in \cite{2017arXiv171010345S}  from
                piecewise-linear activations to arbitrary nonlinear
                smooth ones.
		
                In the one-layer case there exists a simple mapping of
                the continuum GD dynamics of the nonlinear network
                around a minimum to an equivalent linear system with
                an exponential loss by setting
                $x_n' e^{-\frac{1}{2}y_n w'^{T} x_n'} = x_n
                \sqrt{f'(w^T x_n)^2 - f''(w^T x_n)} e^{-\frac{1}{2}y_n
                  w^T x_n}$. The exact mapping of Hessians implies
                again a {\it differentiable conjugacy} of the two dynamical
                systems.
		
              \item If we add a single linear layer on top of
                the one we just considered, i.e.
                $W^2 \sigma(W^1\cdot x_n)$, the second derivative
                becomes
		\begin{equation}
		\label{eq:second_derivative}
		\frac{\partial^{2} f(W;x_n)}{\partial W^k \partial
                 W^{k'} } = 
                \delta_{1 k}\delta_{1 k'}x_n x_n W^2 \cdot \sigma''(W^1\cdot x_n) + 
                \left[\delta_{2 k}\delta_{1 k'}x_n + \delta_{2 k'}\delta_{1 k}x_n \right]\sigma'(W^1\cdot x_n).
		\end{equation}
		 The second
                term here cannot be written as a simple outer product
                of a vector, hence there is no guarantee that the
                Hessian has only $N$ distinct eigenvalues. This
                naturally extends to the case with more
                layers. Indeed, simple numerical checks show that this
                bound is generically broken.
		
		\end{enumerate}	
		
                From the second example it is clear that when we
                consider derivatives across layers, the Hessian of a
                deep network with an exponential loss around a minimum
                for $N$ training examples has more eigenvalues than a
                one layer linear model with the same loss. Adding a
                regularization term $P(W_k) = \lambda |W_k|^2$ helps
                with making all the eigenvalues positive, but does not
                change the number of distinct eigenvalues. Hence,
                unlike the square loss, there does not exist a linear
                model with a single layer which is differentiably
                conjugate to the dynamical system of the deep network
                around a minimum. Nonetheless, after adding an
                arbitrarily small regularization it is possible to
                construct a linear network with an equal number of
                positive eigenvalues. Thus the dynamical system of a deep nonlinear
                network with arbitrarily small regularization
                parameter $\lambda$ around a minimum of gradient
                descent is {\it topologically conjugate} to that of a
                regularized linear network.

                It is natural to ask whether we can strengthen this
                statement into differentiable conjugacy in some
                way. With this in mind, let us consider a deep linear
                network, which also converges in general to the
                minimum norm solution. We have
                $f(W; x_n) = W_{K} W_{K-1} \cdots W_2 W_1 x_n$ for a
                network with $K$ layers. Without loss of generality,
                consider the two-layer case, for which the Hessian is
\begin{equation}
H_{lin} = \sum_n \left[(\delta_{k1} W_2 + \delta_{k2}W_1)(\delta_{k'1} W_2 + \delta_{k'2}W_1)x_n x_n^T - y_n(\delta_{k1}\delta_{k'2}x_n + \delta_{k'1}\delta_{k2}x_n)\right]e^{-y_n W_2 W_1 x_n}.
\end{equation}
This expression clearly cannot be written as a sum of outer products
of $N$ vectors, hence we expect it to have in principle more than $N$
distinct eigenvalues. This is indeed generically true in simulations.

We would like to compare this to the Hessian of a nonlinear deep
network
\begin{equation*}
  H^{nl} = \sum_n \left[\frac{\partial{f(W;x_n )}} {{\partial W_k}} \frac{\partial{f(W;x_n )}} {{\partial W_{k'}}} - y_n \frac{\partial^{2} f(W;x_n)}{\partial W_{k} \partial W_{k'} }\right]e^{-y_n f(W;x_n)},
\end{equation*}
where the second term is given by Equation
(\ref{eq:second_derivative}). To simplify the comparison, let us
consider case when the second derivative $f''$ at the same layer
vanishes, which holds true for rectified nonlinearities. The nonlinear
network can be written as
$f(W;x) = x^T W_1 D_1(x) W_2 D_2(x)\cdots D_{K-1}(x)W_K$, where
$D^t(x_n)$ is a diagonal matrix with entries 0 or 1 giving the profile
of the ReLU activations in layer $t$
\cite{DBLP:journals/corr/abs-1711-01530}. While it is straightforward
to match the value of the nonlinear network to a linear one at a point
for each of the training examples, we do not in principle have enough
variables to match the Hessians exactly.  From the discussion in
Section \ref{Hartman-Grobman theorem} we know that we only
have to match the eigenvalues, rather than matrices. In
\cite{DBLP:journals/corr/SagunBL16} the spectrum of the Hessian of a
deep network with cross-entropy loss was studied numerically and was
shown to be highly degenerate around a minimum of GD.

Since the Hessians are real symmetric matrices, they are linearly
conjugate to diagonal matrices. Thus we obtain two linear systems
$\dot{x}_i = \mu_i x_i$ and $\dot{y}_i = \nu_i y_i$, where $\mu_i$ and
$\nu_i$ are the eigenvalues of the nonlinear and linear deep networks
respectively.  Adding an arbitrarily small regularization $P(w)$ now
gives hyperbolic dynamics, for which the Hartman-Grobman theorem
applies. Ordering the eigenvalues so that
$\mu_1\geq \mu_2\geq\cdots\mu_D>0$ and similarly for $\nu_i$, we can
construct the conjugacies
\begin{equation}
h_i(x_i) = \textnormal{sgn}(x_i) |x|^{\nu_i / \mu_i}.
\end{equation}
If $\mu_i = \nu_i$, then $h_i$ is a diffeomorphism, otherwise it is a
homeomorphism. If there exists a deep linear network with the same
number of distinct eigenvalues as the nonlinear one at a minimum (up
to the freedom of choosing the regularization parameters), then we
obtain differentiable conjugacy. Otherwise there will exist directions
in weight-space in which the equivalence will hold up to a topological
conjugacy. Whether the number of eigenvalues can be matched remains an
open question.

%
%

%

The results above hold not only for the exponential loss, but also for
the family of losses with exponential tails, for example the logistic
function. Note that technically the statements above work for smooth
nonlinearities, for example $\sigma(x) = x/(1+e^{-x/\epsilon^2})$, but
we expect they should apply to non-smooth dynamical systems in the
Filippov sense \cite{arscott1988differential}.

In particular, the results obtained by \cite{2017arXiv171010345S} and  \cite{2018arXiv180600468G} for
the case of linear networks (both shallow and deep) guarantee a ``linearized'' minimum norm
solution in the neighborhood of $\frac{f}{|f|}$ {\it independently} of
the path taken by gradient descent to reach the neighborhood of
$\frac{f}{|f|}$. In our derivation this is because the convergence is
driven by non zero gradient and it is thus independent of initial
conditions. It is important to note that this situation is unlike the
case of the square loss (see Figure \ref{L2Crossentropy}) where the
dependence on initial conditions means that the norm of the local
linearized solution depends on the overall trajectory of gradient
descent and not only on $W_0$.


\section{Early stopping}
\label{Early stopping}
We discuss here  a slightly different  dynamical system minimizing the same
exponential loss function. The dynamics is  related to gradient descent
with batch normalization. 

Consider the usual loss function
$ L(f(w))=\sum_{n=1}^N e^{- f(W;x_n) y_n }$. We define
$W_k=\rho_k V_k$ for $k=1,\cdots,K$ where $K$ is the number of layer
and $W_k$ is the matrix of weights of layer $k$, $V_k$ is the
normalized matrix of weights at layer $k$. Homogeneity of $f$ implies
$f(W;x)=\prod_{k=1}^K \rho_k \tilde{f}(V_1,\cdots,V_K; x_n)$. We
enforce $||V_k||^2=\sum_{i,j} (V_k)_{i,j}^2= 1$ as constraints (any
constant instead of $1$ is acceptable) in the minimization of $L$ by
penalization controlled by $\lambda$. Note that this penalty is
formally different from a regularization parameter since it enforces
unit norm. Thus we are led to finding $V_k$ and $\rho_k$ for which
$L= \sum_{n=1}^N e^{- f(x_n;w) y_n } + \sum_{k=0}^K\lambda_k
(||V_k||^2 -1)$ is zero. We minimize $L$ with respect to $\rho_k,V_k$
by gradient descent. We obtain for $k=1,\cdots,K$

\begin{equation}
\dot{\rho}_k = \sum_n  \rho_1\cdots\rho_{i-1}\rho_{i+1}\cdots \rho_K  e^{-\prod_{i=1}^K \rho_k
\tilde{f}(V_1,\cdots,V_K; x_n)}\tilde{f} (x_n), 
\label{rhodot}
\end{equation}
\noindent and for each layer $k$
\begin{equation}
\dot{V}_k =(\prod_{i=1}^K \rho_i)\sum_n  e^{-\prod_{i=1}^K \rho_i
\tilde{f}(V_1,\cdots,V_K; x_n)} \frac {\partial \tilde{f}(x_n)}
  {\partial V_k} - 2\lambda_k V_k=B_k - 2 \lambda_k V_k
\label{minimumnorm}
\end{equation}
\noindent where
$(\prod_{i=1}^K \rho_i)\sum_n  e^{-\prod_{i=1}^K \rho_i
\tilde{f}(V_1,\cdots,V_K; x_n)} \frac {\partial \tilde{f}(x_n)}
  {\partial V_k}= B_k$.

Observe (see next section) that $\dot{\rho}_k>0$, decreasing to zero for $t \to
\infty$. Also $\lim_{t \to \infty}B_k(t)=0$ from the results in Section \ref{Weightgrowth}.
Note that, since $\frac {\partial
 ||V_k||^2}{\partial t}=2V_k \dot{V_k}$, Equation \ref{minimumnorm} implies
\begin{equation}
\frac {\partial
  ||V_k||^2}{\partial t}=\sum_{i,j}(V_k)_{i,j}(\dot{V}_k)_{i,j} =(\prod_{k=1}^K \rho_k)\sum_n  e^{-\prod_{i=1}^K \rho_i
\tilde{f}(x_n)} \tilde{f}(x_n)- 2 \lambda_k (V_k)^2.
\label{unitnorm}
\end{equation}

 Equation \ref{unitnorm} can be rewritten as 
\begin{equation}
\frac {\partial
  z^k}{\partial t}= C^k(z,\cdots)- 2 \lambda_k z^k
\end{equation}
\noindent with $C^k(z) >0$ decreasing to zero for increasing $t$. When
$C^k=2 \lambda_k$ the equilibrium is reached and $V_k$ has unit norm.

In the approach of this section the values of the $\lambda_k$ are set
by $\dot{V_k}=0$ which enforces the constraint. This means that the
value of $\lambda_k$ effectively determines $T_0$, the time at which
the change in $z$ stop because $C^k(T_0) = 2 \lambda_k(T_0)$.  {\it Thus
  a finite stopping time $T_0$ follows from the value of 
  $\lambda_k(T_0)$.}  The dynamics around the equilibrium point is
hyperbolic for any $\lambda>0$, allowing the use of the Hartman-Grobman theorem. Note
that the unperturbed dynamics around $V_k(T_0)$ is topologically the
same for $\lambda_k (T_0)$ as well as for $\lambda = 0$.  This
suggests a possible approach to prove that {\it (necessary) early
  stopping is equivalent to regularization}. The argument would claim
that in the absence of the $\lambda_k$ terms the dynamics has to be
stopped after a (possibly very long) time $T_0$ and this is equivalent
to a small finite regularization term.


%

Finally, the Hessian of $L$ wrt $V_k$ tells us about the linearized
dynamics around a minimum where the gradient is zero. The Hessian is

\begin{equation}
\sum_n \left[-\left(\prod_{i=1}^K \rho_i^2\right) \frac {\partial \tilde{f} (V;x_n)}
  {\partial V_k} {\frac {\partial \tilde{f} (V;x_n)}
  {\partial V_{k'}}}^T+ \left(\prod_{i=1}^K \rho_i\right) \frac {\partial^2 \tilde{f} (V;x_n)}
  {\partial V_k \partial V_{k'}} \right] e^{- \prod_{i=1}^K \rho_i
    \tilde{f}(V;x_n)} - 2\lambda\mathbf{I}.
\end{equation}

Thus the Hessian is negative semidefinite for $\lambda_k =0$ for large
times because the absolute value of the first term decreases more
slowly than the second term. However, it is asymptotically negative
definite with any $\lambda_k >0$ and thus also in the limit of
$\lambda_k \to 0$.

Putting together all the observations above we have the following
proposition:

\begin{proposition}
  The linearized  dynamics for the exponential loss is
  hyperbolic for large, finite $t$, describing the dynamics of each
  layer weight matrix near a zero minimum of the loss. The
  Hartman-Grobman theorem implies that near a global asymptotic
  minimum $L=0$, for an arbitrarily large finite $T_0$, the linearized
  flow is topologically equivalent to the nonlinear dynamics induced
  by a deep network. The flow converges to the local maximum margin
  solution asymptotically, independently of
  the trajectory leading to the global minimum.

\end{proposition}

{\bf Remarks}

The intuition behind the equations of this section is that if a
solution for the weights $W_k$ exists such that
$y_n f(W, x_n)>0, \quad \forall n$, then the normalized solution also
separates the data. In this case the loss can be made as small as
desired by increasing $\rho_k$. Among all normalized solutions, GD
selects the one with minimum norm {\it because only the nondegenerate
directions around a minimum -- in which the gradient is not zero --
increase}. The degenerate directions, which do not change, are ``washed out'' by
normalization since the effective norm increases steadily during gradient
descent. This justifies the term $\lambda_k ||V_k||^2$ and its limit
for $\lambda_k \to 0$.

Note that we do not assume linearization in the previous
paragraph. Linearization only enters when we consider the Hessian and
its properties around a minimum. It seems therefore possible to
{\it compare meaningfully the norms of different minima} to predict expected
errors. In particular, consider running gradient descent. Assume that
GD converges to an asymptotic global minimum, around which the
increase in the norm is very slow. Set $\epsilon_{T_0}$ and use
Equations \ref{rhodot}, \ref{minimumnorm}, with $\lambda_k=0$, to
compute the products of the Frobenius norms $(\prod_{i=1}^K \rho_i)$
when $L=\epsilon_{T_0}$. This network norm would be a proxy for the
complexity of the network at the specific minimum, allowing a
comparison of  different minimizers.

\section{Rate of growth of weights}\label{Weightgrowth}
In linear 1-layer networks the dynamics of gradient descent yield
$||w||\sim \log t$ asymptotically. For the validity of the results in
the previous section, we need to show that the weights of a deep
network also diverge at infinity. In general, the $K$ nonlinearly
coupled equations (\ref{rhodot}) are not easily solved
analytically. For simplicity of analysis, let us consider the case of
a single training example $N=1$, as we expect the leading asymptotic
behavior to be independent of $N$. In this regime we have
\begin{equation}
\rho_k\dot{\rho}_k = \tilde{f}(x)\left(\prod_{i=1}^k\rho_i\right) e^{-\prod_{i=1}^K\rho_i \tilde{f}(x)} 
\end{equation} 
Keeping all the layers independent makes it difficult to disentangle
for example the behavior of the product of weights
$\prod_{i=1}^K\rho_i$, as even in the 2-layer case the best we can do
is to change variables to $r^2 = \rho^2_1 + \rho^2_2$ and
$\gamma = e^{\rho_1 \rho^2 \tilde{f}(x)}$, for which we still get the
coupled system
\begin{equation}
\dot{\gamma} = \tilde{f}(x)^2 r^2, \qquad r\dot{r} = 2 \frac{\log\gamma}{\gamma},
\end{equation}
from which reading off the asymptotic behavior is nontrivial. 

\begin{figure*}[h!]\centering
	\includegraphics[width=1.0\textwidth]{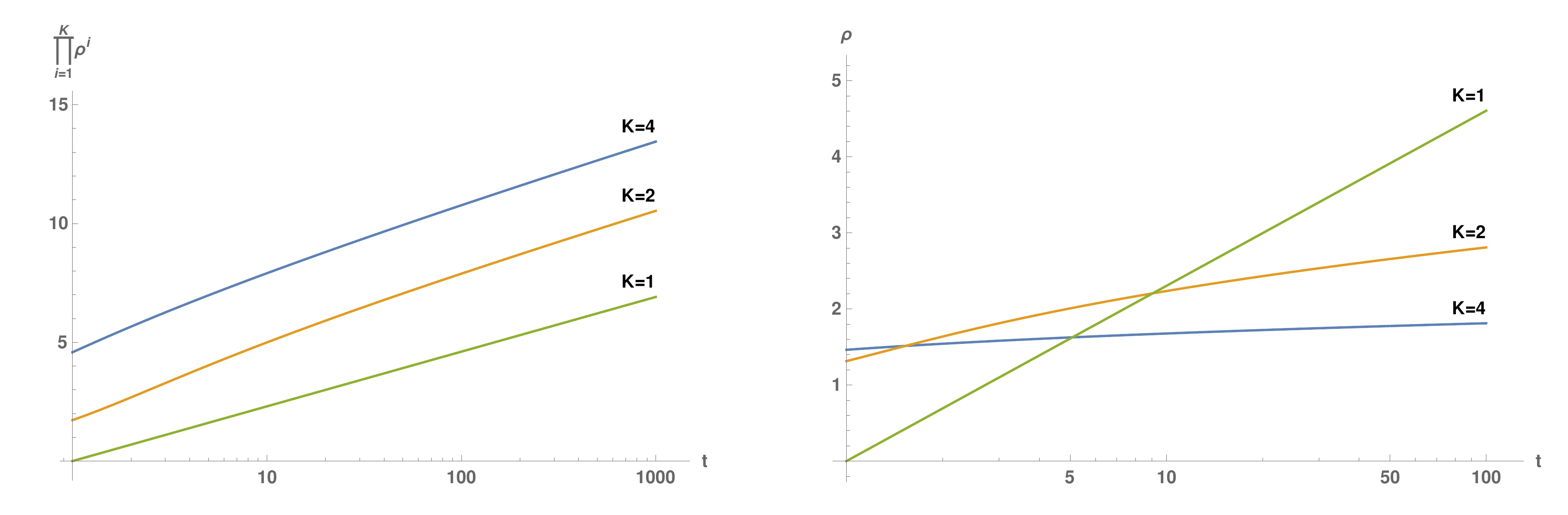}
	\caption{\it The left graph shows how the product of weights
          $\prod_{i=1}^K$ scales as the number of layers grows when
          running gradient descent with an exponential loss. In the
          1-layer case we have $\rho = ||w|| \sim \log t$, whereas for
          deeper networks the product of norms grows faster than
          logarithmically. As we increase the number of
          layers, the individual weights at each layer diverge slower
          than in the 1-layer case, as seen on the right graph.}
	\label{rho_asymptotics}
\end{figure*}

As a simplifying assumption let us consider the case when
$\rho := \rho_1 = \rho_2 = \ldots = \rho_k$. This gives us the single
differential equation
\begin{equation}
\dot{\rho} = \tilde{f}(x) K \rho^{K-1} e^{-\rho_k \tilde{f}(x)}.
\end{equation} 
This implies that for the exponentiated product of weights we have
\begin{equation}
\left(e^{\rho_k \tilde{f}(x)}\right)^{\dot{}} = \tilde{f}(x)^2 K^2 \rho^{2K-2}.
\end{equation}
Changing the variable to $R = e^{\rho_k \tilde{f}(x)}$, we get finally
\begin{equation}
\dot{R} = \tilde{f}(x)^{\frac{2}{K}}K^2 \left(\log R\right)^{2 - \frac{2}{K}}.
\end{equation}
We can now readily check that for $K=1$ we get $R \sim t$, so
$\rho \sim \log t$. It is also immediately clear that for $K>1$ the
product of weights diverges faster than logarithmically. In the case
of $K=2$ we get $R(t) = \textnormal{li}^{-1}(\tilde{f}(x)K^2 t +C)$,
where $\textnormal{li}(z) = \int_0^z dt / \log t$ is the logarithmic
integral function. We show a comparison of the 1-layer and 2-layer
behavior in the left graph in Figure \ref{rho_asymptotics}. For larger
$K$ we get faster divergence, with the limit $K\to\infty$ given by
$R(t) = \mathcal{L}^{-1}(\alpha_\infty t + C)$, where
$\alpha_\infty = \lim_{K\to\infty} \tilde{f}(x)^{\frac{2}{K}}K^2$ and
$\mathcal{L}(z) = \textnormal{li}(z) - \frac{z}{\log z}$.

Interestingly, while the product of weights scales faster than
logarithmically, the weights at each layer diverge slower than in the
linear network case, as can be seen in the right graph in Figure
\ref{rho_asymptotics}.

 {\bf
  Remarks}
\begin{itemize}


\item {\it Cross-entropy loss with Softmax classifier} While the results in the article have been derived for binary classification, they extend to
the case of labels in the set $y_n\in \{1,\ldots,C\}$. In this case we can write the neural network with $K$ layers and rectified nonlinearities $\sigma$ as
\begin{equation}
f(x; W) = \sigma(\sigma(\ldots\sigma(x^T W_1)W_2\ldots W_{K-1})W_K,
\end{equation}
where the last layer $W^K\in\mathbb{R}^{d_K,C}$. In this notation $f(x;W)$ is a $C$-dimensional vector and we can label its' $c$-th component as $f_c$. The cross-entropy loss with the Softmax classifier is then 
\begin{equation}
L = - \sum_{n=1}^N \log \left(\frac{e^{f_{y_n}(x_n;W)}}{\sum_{c=1}^C e^{f_c (x_n;W)}}\right),
\end{equation}
where $f_{y_n}$ is the component of $f(x_n;W)$ corresponding to the correct label for the example $x_n$. The gradient of the loss is then
\begin{equation}
\nabla_W L = \sum_{n=1}^N \sum_{c=1}^C \frac{1}{\sum_{c'=1}^C e^{f_{c'}(x_n; W)-f_c(x_n; W)}}\nabla_W \left(f_c(x_n; W) - f_{y_n}(x_n; W)\right).
\end{equation}
The equivalent assumption of non-linear separability for the cross-entropy loss is that there exists a $W^*$ such that $f_{y_n}(x_n; W^*)- f_c(x_n; W^*) > 0 \ \  \forall n \ \  \forall c\neq y_n$. Using the property of rectified networks $W^T \nabla_W f(x;W) = f(x;W)$, we immediately get that 
$$
W^{*T}\nabla_W L < 0
$$
for any value of $W$. We thus get that as the gradient of the cross-entropy loss $\nabla L \to 0$, the weights $W$ diverge. Rewriting $f_{y_n}(x_n; W)- f_c(x_n; W) = f(x_n;W)$ we see that up to a slightly different normalization (by a sum of exponentials rather than a single exponential) and an additional summation, the dynamics of GD for the cross-entropy loss are those of the exponential loss for binary classification, and as such the results in this article apply also to multi-class classification.

\item {\it GD with weight normalization} Note that the dynamics of
  Equations \ref{rhodot} and \ref{minimumnorm} is different from other
  gradient descent dynamics, though similar.  It represents one of the
  possible approaches for training deep nets on exponential type
  losses: the first approach is to update $W$ and then, in principle
  at least, normalize at the end. The second approach, is similar to
  using weight normalization: GD implements the dynamics of
  $\tilde{w}$. The third approach uses the dynamics corresponding to a
  penalization term enforcing unit Frobenius norm of the weight matrix
  $V_k$.

      \item {\it Non separable case} Consider the linear network in the
        exponential loss  case.
        There will be a finite $w$ for which the gradient is zero. The
        question is whether this is similar to the regularization case
        or not, that is whether {\it misclassification regularizes}.

	Let us look at a linear example:
	
	\begin{equation}
	\dot{w} = F(w) = - \nabla_{w} L(w)= \sum_{n=1}^n x_n^T e^{- x_n^T w} 
	\end{equation}
	
	\noindent in which we assume that there is one classification
        error (say for $n=1$), meaning that the term $e^{- x_1^T w}$
        grows exponentially with $w$. Let us also assume that gradient
        descent converges to $w^*$. This implies that
        $\sum_{n=2}^n x_n^T e^{- x_n^T w^*}= -x_1^T e^{- x_1^T w^*}$:
        for $w^*$ the gradient is zero and $\dot{w}=0$. Is this a
        hyperbolic equilibrium?  Let us look at a very simple $1D$,
        $n=2$ case: 
	
	\begin{equation}
	\dot{w}= - x_1 e^{x_1 w^*}  + x_2 e^{-x_2 w^*}
	\end{equation}
	
	\noindent If $x_2>x_1$ then $\dot{w}=0$ for $ e^{(x_1+x_2)w^*}=\frac{x_2}{x_1}$
	which implies $w^*= \frac{\log(\frac{x_2}{x_1})}{x_1+x_2}$. This is clearly a hyperbolic equilibrium point, since we have
	
	\begin{equation}
	\nabla_w F(w) = -x_1^2 e^{x_1 w^*}  - x_2^2 e^{-x_2 w^*} < 0,
	\end{equation}
	\noindent so the single eigenvalue in this case has no zero real part.
	
	In general, if there are only a small number of classification errors, one
	expects a similar situation for some of the components. {\it
          Differently from the regularization case, misclassification
          errors do not ``regularize'' all components of $w$ but only
          the ones in the span of the misclassified examples}.
\item {\it Learning rate and discretization} In the paper we have
  neglected the time dependence of the learning rate in GD because we
  considered the associated continuous dynamical systems. A
  time-dependent learning
  rate is important when the differential equations are discretized. 
This can be seen by considering the
  differential equation 
\begin{equation}
\frac{dx}{dt} + \gamma(t) x = 0
\label{SDS}
\end{equation}

\noindent with solution $x(t) = x_0 e^{- \int \gamma(t) dt}$. The
condition $\int \gamma(t) dt \rightarrow \infty$ corresponds to $\sum
\gamma_n = \infty$. Conditions of this type are needed for asymptotic
convergence to the minimum of the process $x(t)$. Consider now the
``noisy'' case $\frac{dx}{dt} + \gamma(t) (x + \epsilon(t)) = 0$: we
need $\gamma (t) \epsilon(t) \rightarrow 0$ to eliminate the effect the ``noise'' $\epsilon(t)$, implying at least $\gamma_n \rightarrow
0$. The ``noise'' may just consist of discretization ``noise''. 

\end{itemize}

	\small
\newpage
\bibliographystyle{unsrt}

	\bibliography{Boolean}
	\normalsize

\end{document}